\documentclass[10pt,twocolumn,letterpaper]{article}

\usepackage{iccv}
\usepackage{times}
\usepackage{epsfig}
\usepackage{graphicx}
\usepackage{amsmath}
\usepackage{amssymb}
\usepackage{tablefootnote}

\usepackage{gensymb} 
\usepackage{colortbl}
\usepackage[utf8]{inputenc}

\usepackage[usenames,dvipsnames]{xcolor}
\definecolor{grey}{rgb}{0.8,0.8,0.8}

\graphicspath{{./images/}}

\usepackage[pagebackref=true,breaklinks=true,letterpaper=true,colorlinks,bookmarks=false]{hyperref}

\definecolor{dgreen}{rgb}{0.0,0.7,0.0} 

\iccvfinalcopy 

\def\iccvPaperID{xxx} 

\ificcvfinal\fi

\makeatletter
\renewcommand*{\@fnsymbol}[1]{\ensuremath{\ifcase#1\or *\or \ddagger\or \dagger\or
   \mathsection\or \mathparagraph\or \|\or **\or \dagger\dagger
   \or \ddagger\ddagger \else\@ctrerr\fi}}
\makeatother

\AfterPreamble{\hypersetup{
pagebackref={true},
breaklinks={true},
letterpaper={true},
colorlinks,
bookmarks={false},
citecolor={red},
linkcolor={dgreen}
}}

\newcommand{\sign}{\mathrm{sign}\,}
\newcommand\AF{\fontsize{11}{11.5}\selectfont}
\begin{document}

\title{FlowNet: Learning Optical Flow with Convolutional Networks}

\author{\AF Philipp Fischer\thanks{Supported by the Deutsche Telekom Stiftung}\ \,\thanks{These authors contributed equally},%
\ \ Alexey Dosovitskiy\addtocounter{footnote}{-1}\footnotemark,%
\ \ Eddy Ilg\addtocounter{footnote}{-1}\footnotemark,%
\ \ Philip Häusser,%
\ \ Caner Hazırbaş,%
\ \ Vladimir Golkov\addtocounter{footnote}{-2}\footnotemark\addtocounter{footnote}{1}\\
\AF \quad\qquad University of Freiburg \hspace{3cm} Technical University of Munich\\
{\tt\small \{fischer,dosovits,ilg\}@cs.uni-freiburg.de,\quad \{haeusser,hazirbas,golkov\}@cs.tum.edu}
\and
\AF Patrick van der Smagt\\
\AF Technical University of Munich\\
{\tt\small smagt@brml.org}
\and
\AF Daniel Cremers\\
\AF Technical University of Munich\\
{\tt\small cremers@tum.de}
\and
\AF Thomas Brox\\
\AF University of Freiburg\\
{\tt\small brox@cs.uni-freiburg.de}
}

\maketitle

\begin{abstract}
Convolutional neural networks (CNNs) have recently been very successful in a variety of computer vision tasks, especially on those linked to recognition.
Optical flow estimation has not been among the tasks where CNNs were successful.
In this paper we construct appropriate CNNs which are capable of solving the optical flow estimation problem as a supervised learning task.
We propose and compare two architectures: a generic architecture and another one including a layer that correlates feature vectors at different image locations.

Since existing ground truth datasets are not sufficiently large to train a CNN, we generate a synthetic Flying Chairs dataset.
We show that networks trained on this unrealistic data still generalize very well to existing datasets such as Sintel and KITTI, achieving competitive accuracy at frame rates of 5 to 10 fps.

\end{abstract}

\section{Introduction}\label{sec:intro}

Convolutional neural networks have become the method of choice in many fields of computer vision.
They are classically applied to classification~\cite{lecun1989backpropagation, Krizhevsky-et-al-12}, but recently presented architectures also allow for per-pixel predictions like semantic segmentation~\cite{Long-et-al-15} or depth estimation from single images~\cite{Eigen-et-al-14}.
In this paper, we propose training CNNs end-to-end to learn predicting the optical flow field from a pair of images.

While optical flow estimation needs precise per-pixel localization, it also requires finding correspondences between two input images.
This involves not only learning image feature representations, but also learning to match them at different locations in the two images.
In this respect, optical flow estimation fundamentally differs from previous applications of CNNs.

\begin{figure}
\begin{center}
\includegraphics[width=1.\linewidth]{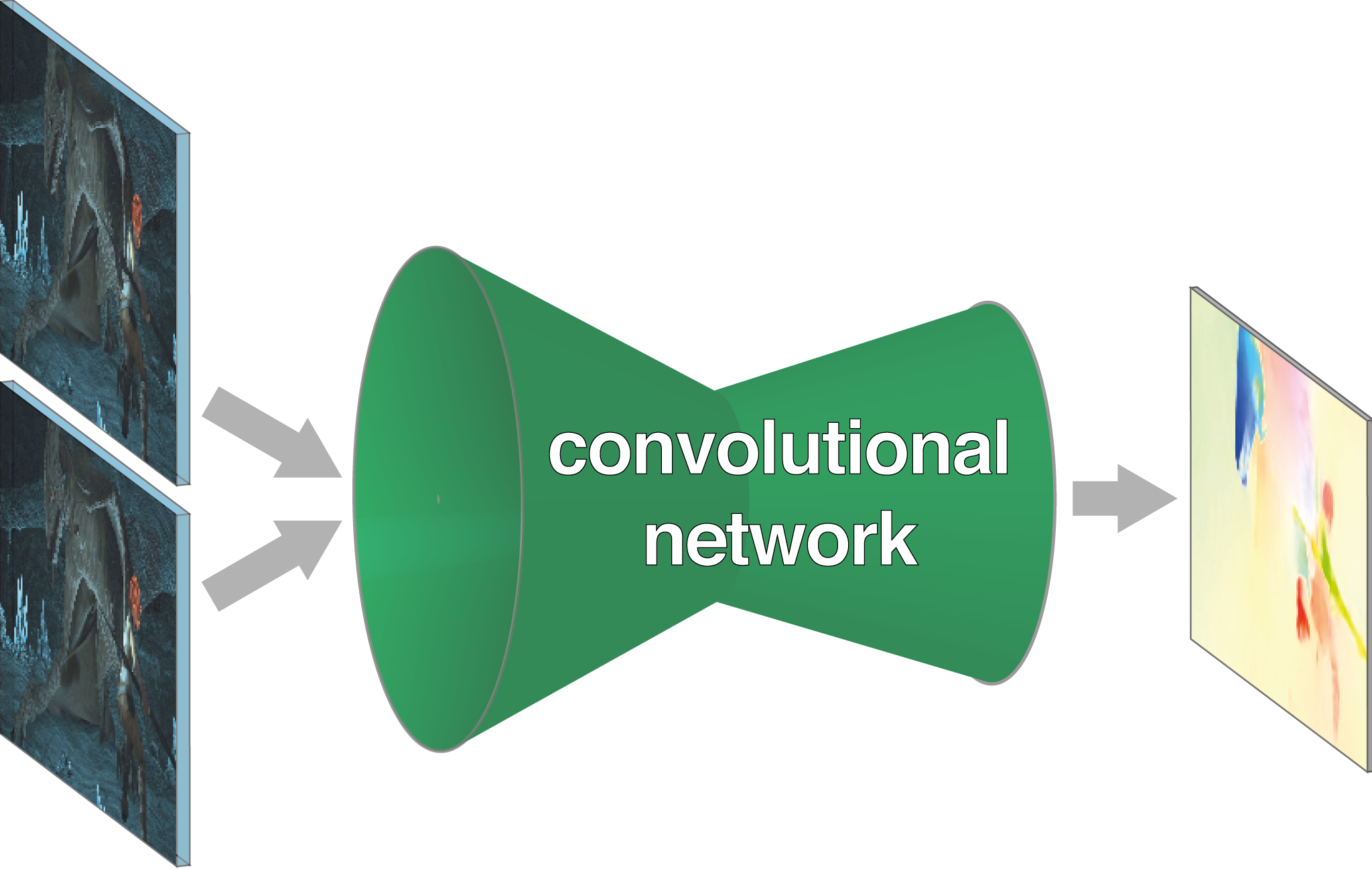}
\end{center}
   \caption{We present neural networks which learn to estimate optical flow, being trained end-to-end. The information is first spatially compressed in a contractive part of the network and then refined in an expanding part.}
\label{fig:long}
\label{fig:onecol}
\end{figure}

Since it was not clear whether this task could be solved with a standard CNN architecture, we additionally developed an architecture with a correlation layer that explicitly provides matching capabilities. 
This architecture is trained end-to-end. 
The idea is to exploit the ability of convolutional networks to learn strong features at multiple levels of scale and abstraction and to help it with finding the actual correspondences based on these features.
The layers on top of the correlation layer learn how to predict flow from these matches.
Surprisingly, helping the network this way is not necessary and even the raw network can learn to predict optical flow with competitive accuracy.

Training such a network to predict generic optical flow requires a sufficiently large training set.
Although data augmentation does help, the existing optical flow datasets are still too small to train a network on par with state of the art.
Getting optical flow ground truth for realistic video material is known to be extremely difficult \cite{Butler-et-al-12}. 
Trading in realism for quantity, we generate a synthetic Flying Chairs dataset which consists of random background images from Flickr on which we overlay segmented images of chairs from~\cite{Aubry-et-al-14}.
These data have little in common with the real world, but we can generate arbitrary amounts of samples with custom properties.
CNNs trained on just these data generalize surprisingly well to realistic datasets, even without fine-tuning.

Leveraging an efficient GPU implementation of CNNs, our method is faster than most competitors.
Our networks predict optical flow at up to $10$ image pairs per second on the full resolution of the Sintel dataset,
achieving state-of-the-art accuracy among real-time methods.

\section{Related Work}\label{sec:related}
\paragraph{Optical Flow.}
Variational approaches have dominated optical flow estimation since the work of Horn and Schunck~\cite{Horn-Schunck-81}.
Many improvements have been introduced \cite{MP98,BBPW04,Wedel-et-al-iccv09}.
The recent focus was on large displacements, and combinatorial matching has been integrated into the variational approach~\cite{Brox-Malik-11,Weinzaepfelet-al-13}.
The work of \cite{Weinzaepfelet-al-13} termed DeepMatching and DeepFlow is related to our work in that feature information is aggregated from fine to coarse using sparse convolutions and max-pooling. However, it does not perform any learning and all parameters are set manually.
The successive work of \cite{Revaud-et-al-15} termed EpicFlow has put even more emphasis on the quality of sparse matching as the matches from \cite{Weinzaepfelet-al-13} are merely interpolated to dense flow fields while respecting image boundaries.
We only use a variational approach for optional refinement of the flow field predicted by the convolutional net and do not require any handcrafted methods for aggregation, matching and interpolation.

Several authors have applied machine learning techniques to optical flow before. 
Sun~\etal~\cite{Sun-08} study statistics of optical flow and learn regularizers using Gaussian scale mixtures; Rosenbaum~\etal~\cite{Rosenbaum-13} model local statistics of optical flow with Gaussian mixture models. 
Black~\etal~\cite{Black-97} compute principal components of a training set of flow fields.
To predict optical flow they then estimate coefficients of a linear combination of these 'basis flows'.
Other methods train classifiers to select among different inertial estimates \cite{Kennedy-et-al-15} 
or to obtain occlusion probabilities \cite{Leordeanu-et-al-13}.

There has been work on unsupervised learning of disparity or motion between frames of videos using neural network models.
These methods typically use multiplicative interactions to model relations between a pair of images.
Disparities and optical flow can then be inferred from the latent variables.
Taylor~\etal~\cite{Taylor-10} approach the task with factored gated restricted Boltzmann machines.
Konda and Memisevic~\cite{Konda-Memisevic-13} use a special autoencoder called `synchrony autoencoder'.
While these approaches work well in a controlled setup and learn features useful for activity recognition in videos,
they are not competitive with classical methods on realistic videos.

\section{Network Architectures}\label{sec:netarchs}
\begin{figure*}
\begin{center}
\begin{tabular}{cc|c}
\includegraphics[width=\linewidth]{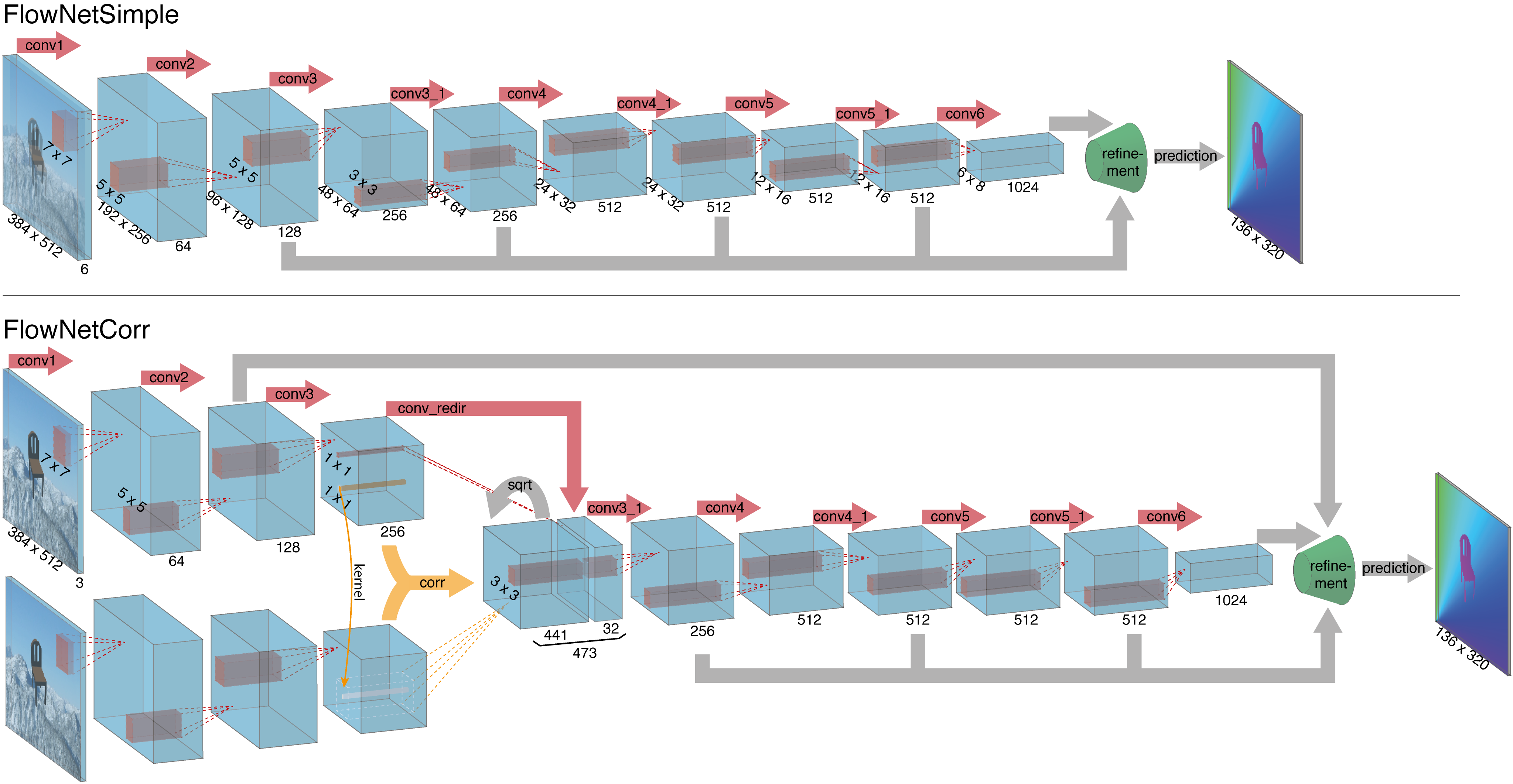}
\end{tabular}
\end{center}
   \caption{The two network architectures: FlowNetSimple (top) and FlowNetCorr (bottom).}
\label{fig:nets}
\end{figure*}

\paragraph{Convolutional Networks.}
Convolutional neural networks trained with backpropagation~\cite{lecun1989backpropagation} have 
recently been shown to perform well on large-scale image classification by Krizhevsky~\etal~\cite{Krizhevsky-et-al-12}.
This gave the beginning to a surge of works on applying CNNs to various computer vision tasks.

While there has been no work on estimating optical flow with CNNs, there has been research on matching with neural networks.
Fischer~\etal~\cite{Fischer-14} extract feature representations from CNNs trained in supervised or unsupervised manner and match these features based on Euclidean distance. 
Zbontar and LeCun~\cite{Zbontar-14} train a CNN with a Siamese architecture to predict similarity of image patches.
A drastic difference of these methods to our approach is that they are patch based and leave the spatial aggregation to post-processing, whereas the networks in this paper directly predict complete flow fields. 

Recent applications of CNNs include semantic segmentation~\cite{Farabet-13, Girshick-et-al-14, Hariharan-et-al-15, Long-et-al-15}, depth prediction~\cite{Eigen-et-al-14}, keypoint prediction~\cite{Hariharan-et-al-15} and edge detection~\cite{Ganin-et-al-14}.
These tasks are similar to optical flow estimation in that they involve per-pixel predictions.
Since our architectures are largely inspired by the recent progress in these per-pixel prediction tasks, we briefly review different approaches.

The simplest solution is to apply a conventional CNN in a `sliding window' fashion, hence computing a single prediction (e.g. class label) for each input image patch~\cite{Ciresan-et-al-12, Farabet-13}.
This works well in many situations, but has drawbacks: high computational costs (even with optimized implementations involving re-usage of intermediate feature maps) and per-patch nature, disallowing to account for global output properties, for example sharp edges. 
Another simple approach~\cite{Hariharan-et-al-15} is to upsample all feature maps to the desired full resolution and stack them together, resulting in a concatenated per-pixel feature vector that can be used to predict the value of interest.

Eigen~\etal~\cite{Eigen-et-al-14} refine a coarse depth map by training an additional network which gets as inputs the coarse prediction and the input image.
Long~\etal~\cite{Long-et-al-15} and Dosovitskiy~\etal~\cite{Dosovitskiy-15} iteratively refine the coarse feature maps with the use of `upconvolutional' layers~\footnote{These layers are often named 'deconvolutional', although the operation they perform is technically convolution, not deconvolution} .
Our approach integrates ideas from both works.
Unlike Long~\etal, we `upconvolve' not just the coarse prediction, but the whole coarse feature maps, allowing to transfer more high-level information to the fine prediction.
Unlike Dosovitskiy~\etal, we concatenate the `upconvolution' results with the features from the `contractive' part of the network.

Convolutional neural networks are known to be very good at learning input--output relations given enough labeled data.
We therefore take an end-to-end learning approach to predicting optical flow: given a dataset consisting of image pairs and ground truth flows, we train a network to predict the $x$--$y$ flow fields directly from the images.
But what is a good architecture for this purpose?

A simple choice is to stack both input images together and feed them through a rather generic network, allowing the network to decide itself how to process the image pair to extract the motion information.
This is illustrated in Fig.~\ref{fig:nets} (top).
We call this architecture consisting only of convolutional layers `FlowNetSimple'.

In principle, if this network is large enough, it could learn to predict optical flow.
However, we can never be sure that a local gradient optimization like stochastic gradient descent can get the network to this point.
Therefore, it could be beneficial to hand-design an architecture which is less generic, but may perform better with the given data and optimization techniques.

A straightforward step is to create two separate, yet identical processing streams for the two images and to combine them at a later stage as shown in Fig.~\ref{fig:nets} (bottom).
With this architecture the network is constrained to first produce meaningful representations of the two images separately and then combine them on a higher level.
This roughly resembles the standard matching approach when one first extracts features from patches of both images and then compares those feature vectors.
However, given feature representations of two images, how would the network find correspondences?

To aid the network in this matching process, we introduce a `correlation layer' that performs multiplicative patch comparisons between two feature maps.
An illustration of the network architecture `FlowNetCorr' containing this layer is shown in Fig.~\ref{fig:nets} (bottom).
Given two multi-channel feature maps $\mathbf{f}_1, \mathbf{f}_2 : \mathbb{R}^2 \to \mathbb{R}^c$, with $w$, $h$, and $c$ being their width, height and number of channels,
our correlation layer lets the network compare each patch from $\mathbf{f}_1$ with each path from $\mathbf{f}_2$.

For now we consider only a single comparison of two patches.
The 'correlation' of two patches centered at $\mathbf{x}_1$ in the first map and $\mathbf{x}_2$ in the second map is then defined as
\begin{equation}
  c(\mathbf{x}_1,\mathbf{x}_2) = \mkern-18mu \sum_{\mathbf{o} \in [-k,k] \times [-k,k] } \mkern-36mu\langle\mathbf{f}_1(\mathbf{x}_1+\mathbf{o}),\mathbf{f}_2(\mathbf{x}_2+\mathbf{o})\rangle
  \label{eq:patch_correlation}
\end{equation}
for a square patch of size $K:= 2k+1$.
Note that Eq.~\ref{eq:patch_correlation} is identical to one step of a convolution in neural networks, but instead of convolving data with a filter, it convolves data with other data.
For this reason, it has no trainable weights.

Computing $c(\mathbf{x}_1,\mathbf{x}_2)$ involves $c \cdot K^2$ multiplications.
Comparing all patch combinations involves $w^2\cdot h^2$ such computations, yields a large result and makes efficient forward and backward passes intractable.
Thus, for computational reasons we limit the maximum displacement for comparisons and also introduce striding in both feature maps.

Given a maximum displacement $d$, for each location $\mathbf{x}_1$ we compute correlations $c(\mathbf{x}_1,\mathbf{x}_2)$ only in a neighborhood of size $D:=2d+1$, by limiting the range of $\mathbf{x}_2$.
We use strides $s_1$ and $s_2$, to quantize $\mathbf{x}_1$ globally and to quantize $\mathbf{x}_2$ within the neighborhood centered around $\mathbf{x}_1$.

In theory, the result produced by the correlation is four-dimensional: for every combination of two 2D positions we obtain a correlation value, i.e. the scalar product of the two vectors which contain the values of the cropped patches respectively.
In practice we organize the relative displacements in channels.
This means we obtain an output of size $(w \times h \times D^2)$.
For the backward pass we implemented the derivatives with respect to each bottom blob accordingly.
\begin{figure}[t]
\begin{center}
\includegraphics[width=1.\linewidth]{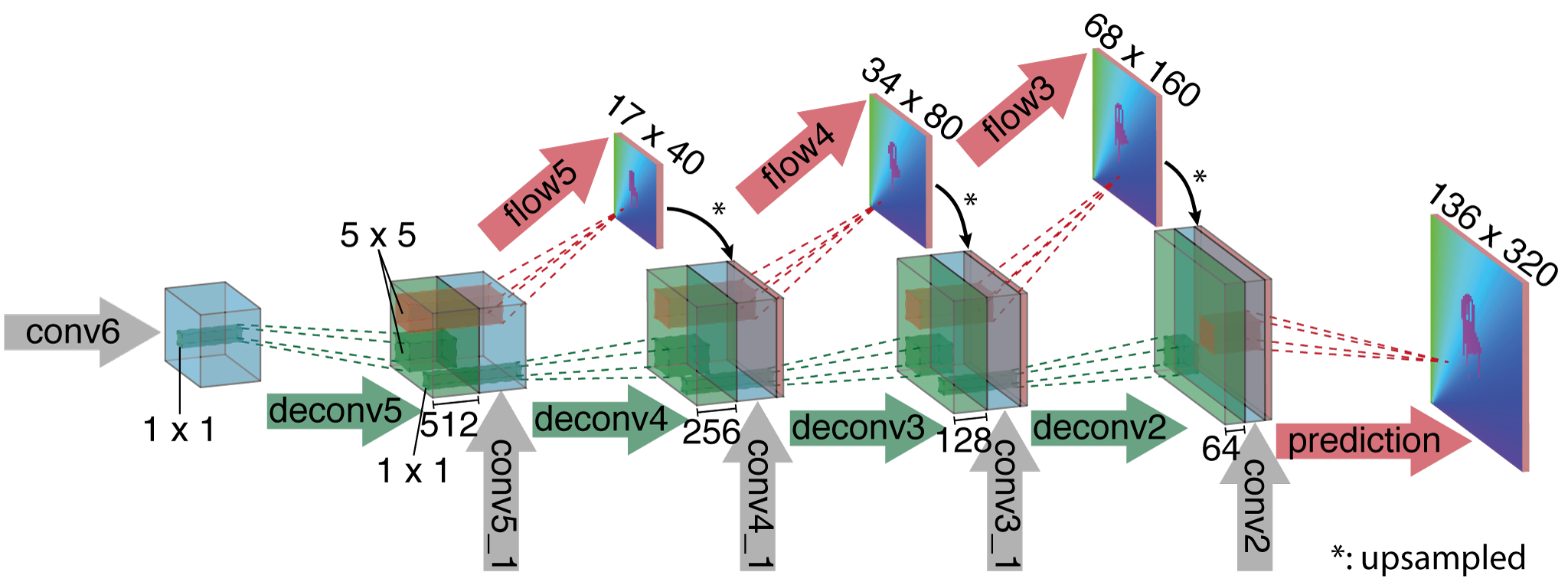}
\end{center}
   \caption{Refinement of the coarse feature maps to the high resolution prediction.}
\label{fig:refinement}
\end{figure}

\paragraph{Refinement.}
CNNs are good at extracting high-level abstract features of images, by interleaving convolutional layers and pooling, i.e. spatially shrinking the feature maps.
Pooling is necessary to make network training computationally feasible and, more fundamentally, to allow aggregation of information over large areas of the input images.
However, pooling results in reduced resolution, so in order to provide dense per-pixel predictions we need a way to refine the coarse pooled representation.

Our approach to this refinement is depicted in Figure~\ref{fig:refinement}.
The main ingredient are `upconvolutional' layers, consisting of unpooling (extending the feature maps, as opposed to pooling) and a convolution.
Such layers have been used previously~\cite{Zeiler-11, Zeiler-14, Goodfellow-14, Long-et-al-15, Dosovitskiy-15}.
To perform the refinement, we apply the `upconvolution' to feature maps, and concatenate it with corresponding feature maps from the 'contractive' part of the network and an upsampled coarser flow prediction (if available).
This way we preserve both the high-level information passed from coarser feature maps and fine local information provided in lower layer feature maps.
Each step increases the resolution twice. 
We repeat this $4$ times, resulting in a predicted flow for which the resolution is still $4$ times smaller than the input.

We discover that further refinement from this resolution does not significantly improve the results, compared to a computationally less expensive bilinear upsampling to full image resolution. 
The result of this bilinear upsampling is the final flow predicted by the network.

\begin{figure}[t]
\begin{center}
\setlength{\tabcolsep}{0.03cm}
\renewcommand{\arraystretch}{0.5}
  \begin{tabular}{ccc}
  Ground truth & FlowNetS & FlowNetS+v
  \\
  {\includegraphics[width=0.33\linewidth]{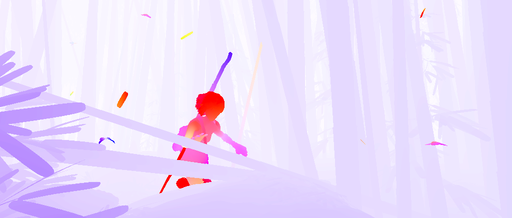}} &
  {\includegraphics[width=0.33\linewidth]{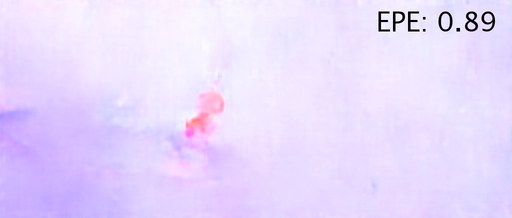}} &
  {\includegraphics[width=0.33\linewidth]{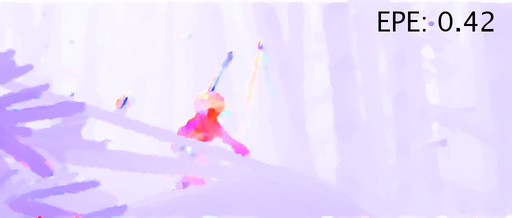}}
  \\
  {\includegraphics[width=0.33\linewidth]{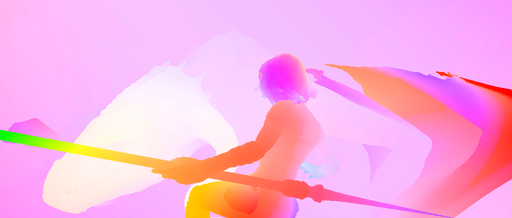}} &
  {\includegraphics[width=0.33\linewidth]{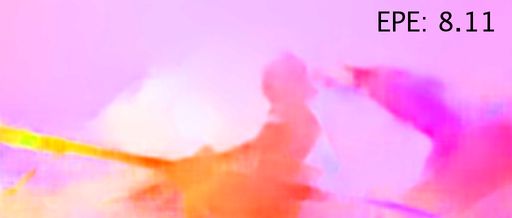}} &
  {\includegraphics[width=0.33\linewidth]{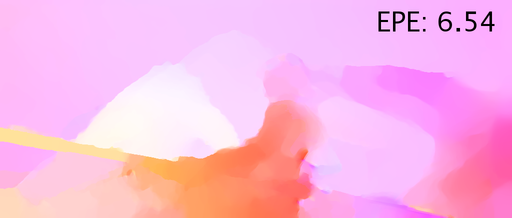}}
  \end{tabular}
\end{center}
   \caption{The effect of variational refinement. 
   In case of small motions (first row) the predicted flow is changed dramatically.
   For larger motions (second row), big errors are not corrected, but the flow field is smoothed, resulting in lower EPE. }
\label{fig:var_refinement_examples}
\end{figure}

In an alternative scheme, instead of bilinear upsampling we use the variational approach from~\cite{Brox-Malik-11} without the matching term:
we start at the $4$ times downsampled resolution and then 
use the coarse to fine scheme with $20$ iterations to bring the flow field to the full resolution. 
Finally, we run $5$ more iterations at the full image resolution.
We additionally compute image boundaries with the approach from~\cite{Leordeanu-et-al-12} and 
respect the detected boundaries by replacing the 
smoothness coefficient by $\alpha=\exp(-\lambda b(x,y) ^ \kappa$), where $b(x,y)$ 
denotes the thin boundary strength resampled at the respective scale and between pixels. 
This upscaling method is more computationally expensive than simple bilinear upsampling, but adds the benefits of 
variational methods to obtain smooth and subpixel-accurate flow fields. In the following, 
we denote the results obtained by this variational refinement with a `+v' suffix.
An example of variational refinement can be seen in Fig.~\ref{fig:var_refinement_examples}.
\begin{table}[b]
\begin{tabular}{l|rrr}
              & Frame  & Frames with  & Ground truth       \\
              & pairs  & ground truth & density per frame  \\
\hline
Middlebury    & 72     & 8            & 100\%              \\
KITTI         & 194    & 194          & $\backsim$50\%      \\
Sintel 	      & 1,041  & 1,041        & 100\%              \\
Flying Chairs & 22,872 & 22,872       & 100\%              \\
\end{tabular}
\vspace{0.1em}
\caption{Size of already available datasets and the proposed Flying Chairs dataset.}
\label{tab:dataset_comparison}
\end{table}

\section{Training Data}\label{sec:traindata}
Unlike traditional approaches, neural networks require data with ground truth not only for optimizing several parameters, but to learn to perform the task from scratch.
In general, obtaining such ground truth is hard, because true pixel correspondences for real world scenes cannot easily be determined.
An overview of the available datasets is given in Table \ref{tab:dataset_comparison}.

\subsection{Existing Datasets}\label{sec:dataset}
The Middlebury dataset~\cite{Baker-et-al-09} contains only 8 image pairs for training, with ground truth flows generated using four different techniques.
Displacements are very small, typically below $10$ pixels.

The KITTI dataset~\cite{Geiger-et-al-13} is larger (194 training image pairs) and includes large displacements, but contains only a very special motion type.
The ground truth is obtained from real world scenes by simultaneously recording the scenes with a camera and a 3D laser scanner.
This assumes that the scene is rigid and that the motion stems from a moving observer.
Moreover, motion of distant objects, such as the sky, cannot be captured, resulting in sparse optical flow ground truth.

The MPI Sintel~\cite{Butler-et-al-12} dataset obtains ground truth from rendered artificial scenes with special attention to realistic image properties. 
Two versions are provided: the Final version contains motion blur and atmospheric effects, such as fog, while the Clean version does not include these effects. 
Sintel is the largest dataset available (1,041 training image pairs for each version) and provides dense ground truth for small and large displacement magnitudes.

\begin{figure*}
\begin{center}
        {\includegraphics[width = 2.6cm]{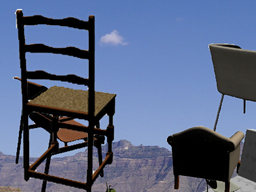}}
        {\includegraphics[width = 2.6cm]{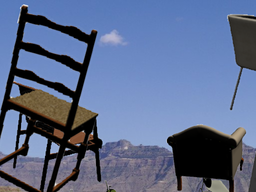}}
        {\includegraphics[width = 2.6cm]{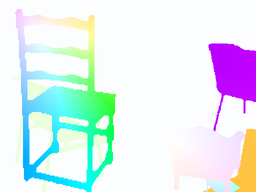}}
        {\includegraphics[width = 2.6cm]{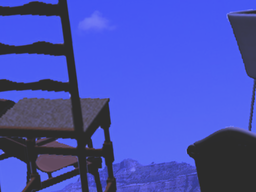}}
        {\includegraphics[width = 2.6cm]{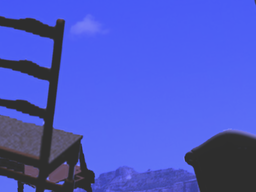}}
        {\includegraphics[width = 2.6cm]{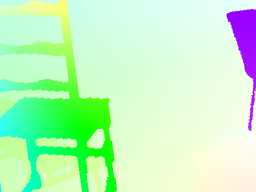}}
        \\
      {\includegraphics[width = 2.6cm]{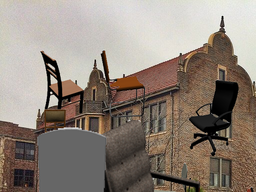}}
      {\includegraphics[width = 2.6cm]{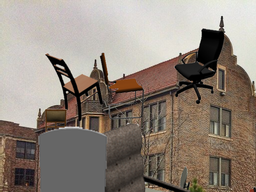}}
      {\includegraphics[width = 2.6cm]{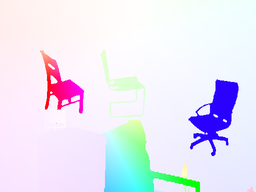}}
      {\includegraphics[width = 2.6cm]{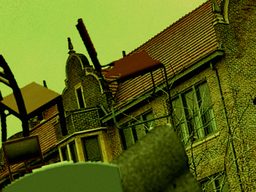}}
      {\includegraphics[width = 2.6cm]{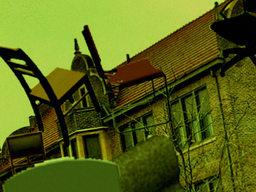}}
      {\includegraphics[width = 2.6cm]{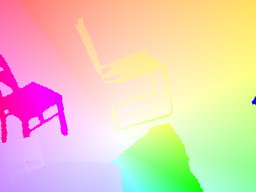}}
        
\end{center}
        \caption{Two examples from the Flying Chairs dataset. Generated image pair and color coded flow field (first three columns), augmented image pair and corresponding color coded flow field respectively (last three columns).}
\label{fig:FlyingChairs}
\end{figure*}

\subsection{Flying Chairs}\label{sec:chairs}
The Sintel dataset is still too small to train large CNNs.
To provide enough training data, we create a simple synthetic dataset, which we name Flying Chairs,
by applying affine transformations to images collected from Flickr and a publicly available rendered set of 3D chair models~\cite{Aubry-et-al-14}.
We retrieve $964$ images from Flickr\footnote{Non-commercial public license. We use the code framework by Hays and Efros~\cite{Hays-Efros-08}} with a resolution of $1,024\times 768$ from the categories `city' ($321$), `landscape' ($129$) and `mountain' ($514$). We cut the images into 4 quadrants and use the resulting $512 \times 384$ image crops as background.
As foreground objects we add images of multiple chairs from ~\cite{Aubry-et-al-14} to the background.
From the original dataset we remove very similar chairs, resulting in $809$ chair types and $62$ views per chair available.
Examples are shown in Figure~\ref{fig:FlyingChairs}.

To generate motion, we randomly sample affine transformation parameters for the background and the chairs.
The chairs' transformations are relative to the background transformation, which can be interpreted as both the camera and the objects moving.
Using the transformation parameters we render the second image, the optical flow and occlusion regions.

All parameters for each image pair (number, types, sizes and initial positions of the chairs; transformation parameters) are randomly sampled.
We adjust the random distributions of these parameters in such a way that the resulting displacement histogram is similar to the one from Sintel (details can be found in the supplementary material).
Using this procedure, we generate a dataset with 22,872 image pairs and flow fields (we re-use each background image multiple times).
Note that this size is chosen arbitrarily and could be larger in principle.

\subsection{Data Augmentation}\label{sec:augmentation}
A widely used strategy to improve generalization of neural networks is data augmentation~\cite{Krizhevsky-et-al-12, Eigen-et-al-14}.
Even though the Flying Chairs dataset is fairly large, we find that using augmentations is crucial to avoid overfitting.
We perform augmentation online during network training.
The augmentations we use include geometric transformations: \emph{translation}, \emph{rotation} and \emph{scaling},
as well as additive \emph{Gaussian noise} and changes in \emph{brightness}, \emph{contrast}, \emph{gamma}, and \emph{color}.
To be reasonably quick, all these operations are processed on the GPU. Some examples of augmentation are given in Fig.~\ref{fig:FlyingChairs}.

As we want to increase not only the variety of images but also the variety of flow fields, we apply the same strong geometric transformation to both images of a pair, but additionally a smaller relative transformation between the two images.
We adapt the flow field accordingly by applying the per-image augmentations to the flow field from either side.

Specifically we sample \emph{translation} from a the range $[-20\%,\, 20\%]$ of the image width for $x$ and $y$; \emph{rotation} from
$[-17\degree,\, 17\degree]$; \emph{scaling} from $[0.9,\, 2.0]$.
The Gaussian \emph{noise} has a sigma uniformly sampled from $[0,\,0.04]$;
\emph{contrast} is sampled within $[-0.8,\,0.4]$; multiplicative color changes to the RGB channels per image from $[0.5,\,2]$; gamma values from $[0.7,\,1.5]$ and additive brightness changes using Gaussian with a sigma of $0.2$.

\section{Experiments}\label{sec:experiments}
We report the results of our networks on the Sintel, KITTI and Middlebury datasets, as well as on our synthetic Flying Chairs dataset.
We also experiment with fine-tuning of the networks on Sintel data and variational refinement of the predicted flow fields.
Additionally, we report runtimes of our networks, in comparison to other methods.

\subsection{Network and Training Details}
The exact architectures of the networks we train are shown in Fig.~\ref{fig:nets}.
Overall, we try to keep the architectures of different networks consistent: they have nine convolutional layers with stride of $2$ (the simplest form of pooling) in six of them and a ReLU nonlinearity after each layer.
We do not have any fully connected layers, which allows the networks to take images of arbitrary size as input.
Convolutional filter sizes decrease towards deeper layers of networks: $7\times 7$ for the first layer, $5\times 5$ for the following two layers and $3\times 3$ starting from the fourth layer.
The number of feature maps increases in the deeper layers, roughly doubling after each layer with a stride of $2$.
For the correlation layer in FlowNetC we chose the parameters $k=0$, $d=20$, $s_1=1$, $s_2=2$.
As training loss we use the endpoint error~(EPE), which is the standard error measure for optical flow estimation.
It is the Euclidean distance between the predicted flow vector and the ground truth, averaged over all pixels.

For training CNNs we use a modified version of the caffe~\cite{caffe} framework. 
We choose Adam~\cite{Kingma-14} as optimization method because for our task it shows faster convergence than standard stochastic gradient descent with momentum. 
We fix the parameters of Adam as recommended in~\cite{Kingma-14}: $\beta_1 = 0.9$ and $\beta_2 = 0.999$.
Since, in a sense, every pixel is a training sample, we use fairly small mini-batches of $8$ image pairs.
We start with learning rate $\lambda = 1e\!\!-\!\!4$ and then divide it by $2$ every $100$k iterations after the first $300$k.
With FlowNetCorr we observe exploding gradients with $\lambda=1e\!\!-\!\!4$.
To tackle this problem, we start by training with a very low learning rate $\lambda = 1e\!\!-\!\!6$, slowly increase it to reach $\lambda = 1e\!\!-\!\!4$ after $10$k iterations and then follow the schedule just described.

To monitor overfitting during training and fine-tuning, we split
the Flying Chairs dataset into $22,232$ training and $640$ test samples and split
the Sintel training set into $908$ training and $133$ validation pairs.

We found that upscaling the input images during testing may improve the performance. Although the optimal scale depends on the specific dataset, we fixed the scale once for each network for all tasks. For FlowNetS we do not upscale, for FlowNetC we chose a factor of $1.25$.

\paragraph{Fine-tuning.}
The used datasets are very different in terms of object types and motions they include.
A standard solution is to fine-tune the networks on the target datasets.
The KITTI dataset is small and only has sparse flow ground truth. Therefore, we choose to fine-tune on the Sintel training set.
We use images from the Clean and Final versions of Sintel together and fine-tune using a low learning rate $\lambda = 1e\!\!-\!\!6$ for several thousand iterations.
For best performance, after defining the optimal number of iterations using a validation set, we then fine-tune on the whole training set for the same number of iterations.
In tables we denote fine-tuned networks with a `+ft' suffix.

\subsection{Results}
Table~\ref{tbl:results_big} shows the endpoint error (EPE) of our networks and several well-performing methods on public datasets (Sintel, KITTI, Middlebury), as well as on our Flying Chairs dataset. 
Additionally we show runtimes of different methods on Sintel.

The networks trained just on the non-realistic Flying Chairs perform very well on real optical flow datasets, beating for example the well-known LDOF~\cite{Brox-Malik-11} method.
After fine-tuning on Sintel our networks can outperform the competing real-time method EPPM~\cite{Bao-et-al-14} on Sintel Final and KITTI while being twice as fast.

\begin{table*}
\begin{center}
\resizebox{0.93\textwidth}{!}{%
\begin{tabular}{l||cc|cc|cc|cccc|c|cc}
\hline
Method                              &  \multicolumn{2}{c|}{Sintel Clean} & \multicolumn{2}{c|}{Sintel Final} & \multicolumn{2}{c|}{KITTI} & \multicolumn{2}{c}{Middlebury train}  & \multicolumn{2}{c|}{Middlebury test} & Chairs & \multicolumn{2}{c}{Time (sec)}\\
                                    & train	& test & train & test & train & test & AEE & AAE & AEE & AAE & test & CPU & GPU \\
\hline
EpicFlow~\cite{Revaud-et-al-15}     & $2.40$  & $4.12$  & $3.70$  & $6.29$  & $3.47$  & $3.8$  & $0.31$ &   $3.24$ & $0.39$  & $3.55$ & $2.94$ & $16$  & -        \\
DeepFlow~\cite{Weinzaepfelet-al-13} & $3.31$  & $5.38$	& $4.56$  & $7.21$  & $4.58$  & $5.8$  & $0.21$ &   $3.04$ & $0.42$  & $4.22$ & $3.53$ & $17$  & -        \\
EPPM~\cite{Bao-et-al-14}            &      -  & $6.49$  &      -  & $8.38$  & -       & $9.2$  &      - &        - & $0.33$  & $3.36$ &     -  & -     & $0.2$    \\
LDOF~\cite{Brox-Malik-11}           & $4.29$  & $7.56$  & $6.42$  & $9.12$  & $13.73$ & $12.4$ & $0.45$ &   $4.97$ & $0.56$  & $4.55$ & $3.47$ & $65$  & $2.5$    \\
\hline
FlowNetS                            & $4.50$  & $7.42$  & $5.45$  & $8.43$  & $8.26$  & -      & $1.09$ &  $13.28$ & -       & -      & $2.71$ & -     & $0.08$   \\
FlowNetS+v                          & $3.66$  & $6.45$  & $4.76$  & $7.67$  & $6.50$  & -      & $0.33$ &   $3.87$ & -       & -      & $2.86$ & -     & $1.05$   \\
FlowNetS+ft                         & $(3.66)$& $6.96$  & $(4.44)$& $7.76$  & $7.52$  & $9.1$  & $0.98$ &  $15.20$ & -       & -      & $3.04$ & -     & $0.08$   \\
FlowNetS+ft+v                       & $(2.97)$& $6.16$  & $(4.07)$& $7.22$  & $6.07$  & $7.6$  & $0.32$ &   $3.84$ & $0.47$  & $4.58$ & $3.03$ & -     & $1.05$   \\
\arrayrulecolor{grey} \hline \arrayrulecolor{black} 
FlowNetC                            & $4.31$  & $7.28$  & $5.87$  & $8.81$  & $9.35$  & -      & $1.15$ &  $15.64$ & -       & -      & $2.19$ & -     & $0.15$   \\
FlowNetC+v                          & $3.57$  & $6.27$  & $5.25$  & $8.01$  & $7.45$  & -      & $0.34$ &   $3.92$ & -       & -      & $2.61$ & -     & $1.12$   \\
FlowNetC+ft                         & $(3.78)$& $6.85$  & $(5.28)$& $8.51$  & $8.79$  & -      & $0.93$ &  $12.33$ & -       & -      & $2.27$ & -     & $0.15$   \\
FlowNetC+ft+v                       & $(3.20)$& $6.08$  & $(4.83)$& $7.88$  & $7.31$  & -      & $0.33$ &   $3.81$ & $0.50$  & $4.52$ & $2.67$ & -     & $1.12$   \\
\hline
\end{tabular}}
\end{center}
\caption{Average endpoint errors (in pixels) of our networks compared to several well-performing methods on different datasets. The numbers in parentheses are the results of the networks on data they were trained on, and hence are not directly comparable to other results. }\label{tbl:results_big}
\end{table*}

\paragraph{Sintel.}

\begin{figure*}
\begin{center}
\setlength{\tabcolsep}{0.03cm}
\renewcommand{\arraystretch}{0.5}
  \begin{tabular}{ccccc}
  Images & Ground truth & EpicFlow & FlowNetS & FlowNetC
  \\
  {\includegraphics[width=0.2\linewidth]{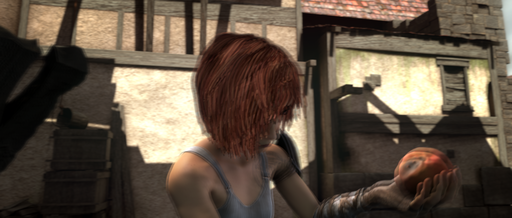}} &
  {\includegraphics[width=0.2\linewidth]{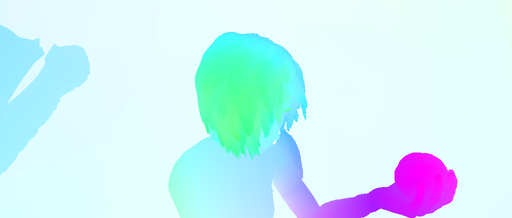}} &
  {\includegraphics[width=0.2\linewidth]{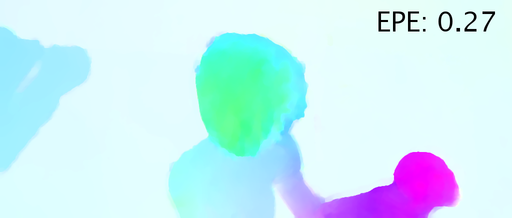}} &
  {\includegraphics[width=0.2\linewidth]{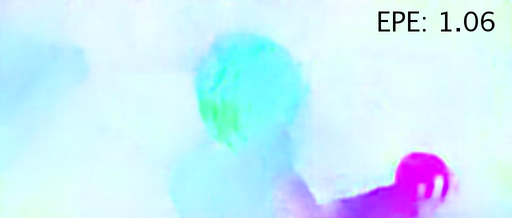}} &
  {\includegraphics[width=0.2\linewidth]{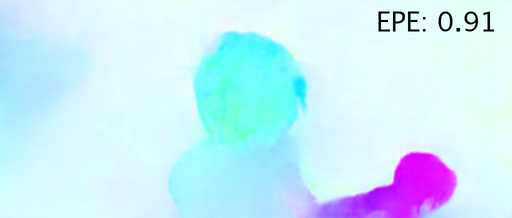}}
  \\
  {\includegraphics[width=0.2\linewidth]{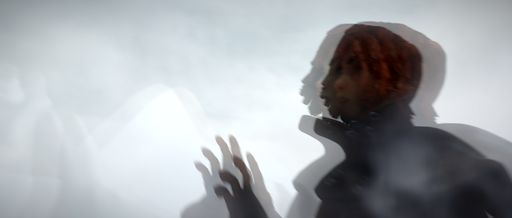}} &
  {\includegraphics[width=0.2\linewidth]{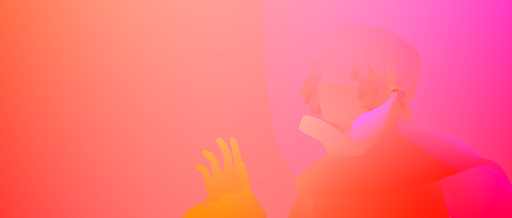}} &
  {\includegraphics[width=0.2\linewidth]{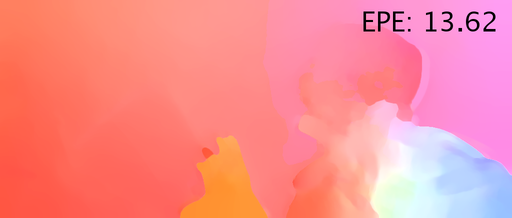}} &
  {\includegraphics[width=0.2\linewidth]{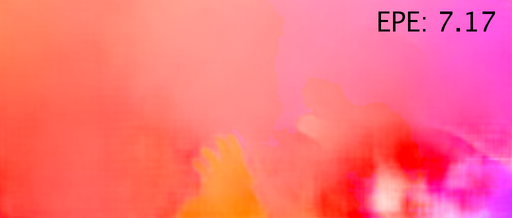}} &
  {\includegraphics[width=0.2\linewidth]{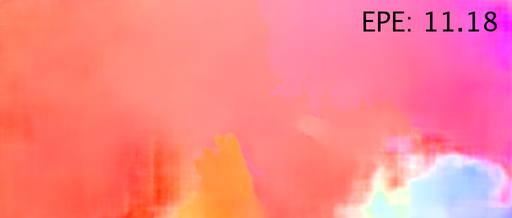}}
  \\
  {\includegraphics[width=0.2\linewidth]{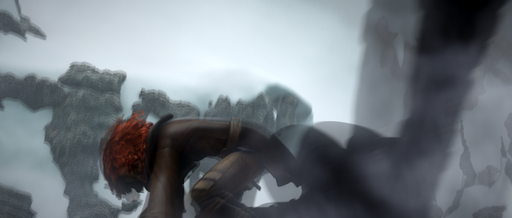}} &
  {\includegraphics[width=0.2\linewidth]{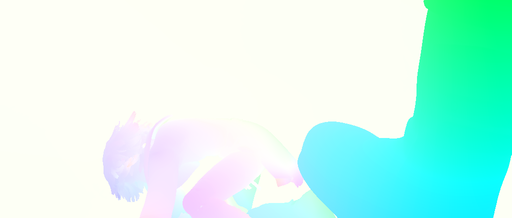}} &
  {\includegraphics[width=0.2\linewidth]{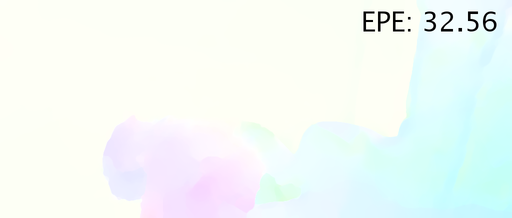}} &
  {\includegraphics[width=0.2\linewidth]{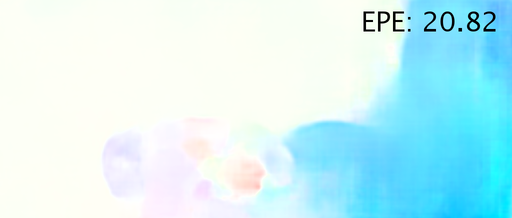}} &
  {\includegraphics[width=0.2\linewidth]{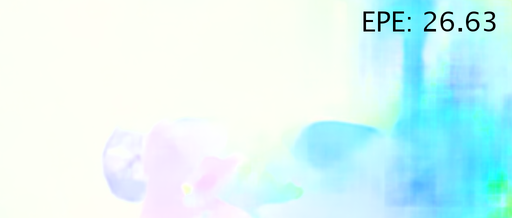}}
  \\
  {\includegraphics[width=0.2\linewidth]{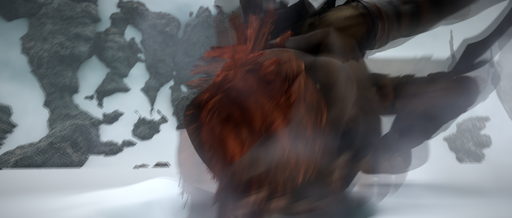}} &
  {\includegraphics[width=0.2\linewidth]{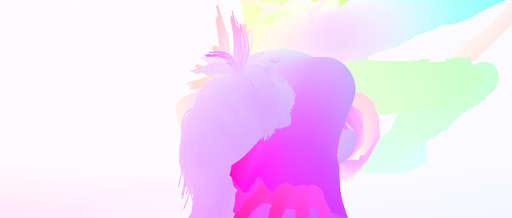}} &
  {\includegraphics[width=0.2\linewidth]{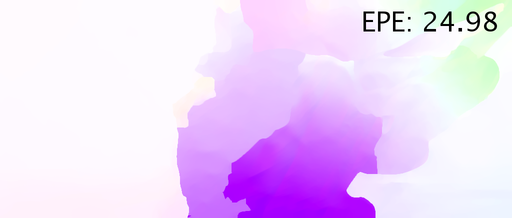}} &
  {\includegraphics[width=0.2\linewidth]{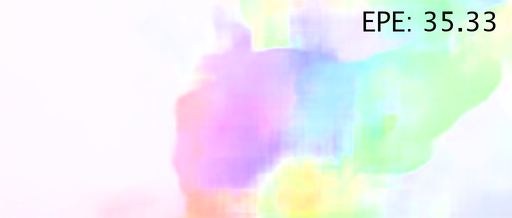}} &
  {\includegraphics[width=0.2\linewidth]{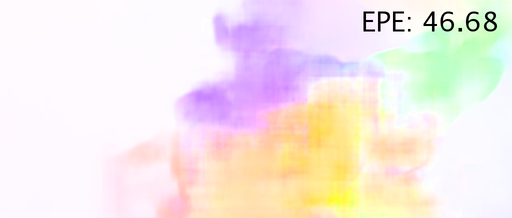}}
  \\
  {\includegraphics[width=0.2\linewidth]{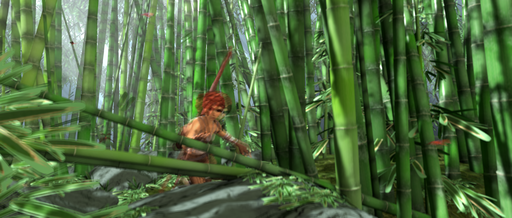}} &
  {\includegraphics[width=0.2\linewidth]{./resources/flow_examples/sintel.train.final_bamboo_2_30/gt.png}} &
  {\includegraphics[width=0.2\linewidth]{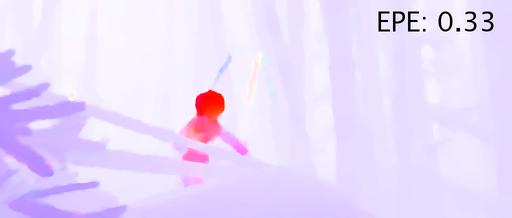}} &
  {\includegraphics[width=0.2\linewidth]{./resources/flow_examples/sintel.train.final_bamboo_2_30/dumbnet_43_chairs_moreds_540000_s=1.png}} &
  {\includegraphics[width=0.2\linewidth]{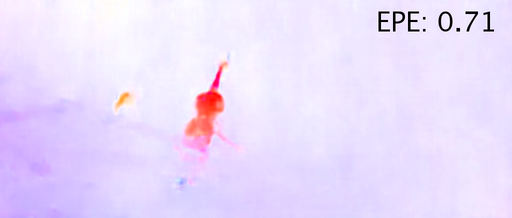}}
  \\
  {\includegraphics[width=0.2\linewidth]{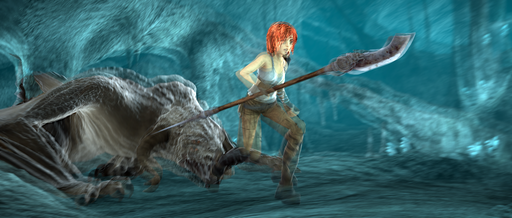}} &
  {\includegraphics[width=0.2\linewidth]{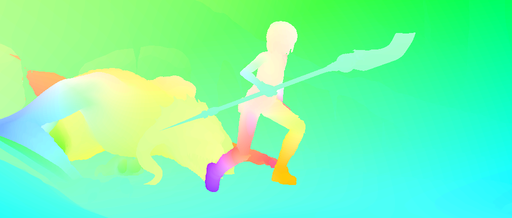}} &
  {\includegraphics[width=0.2\linewidth]{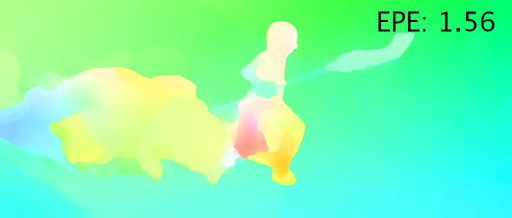}} &
  {\includegraphics[width=0.2\linewidth]{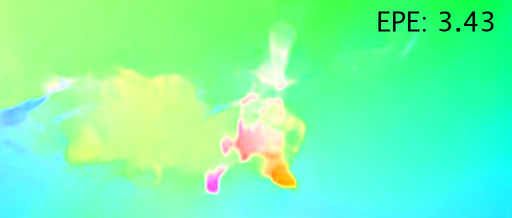}} &
  {\includegraphics[width=0.2\linewidth]{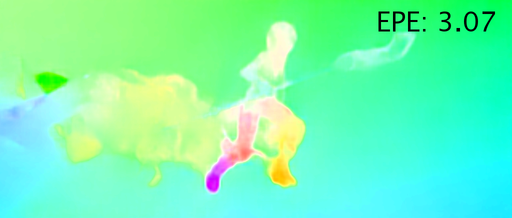}}
  \\
  {\includegraphics[width=0.2\linewidth]{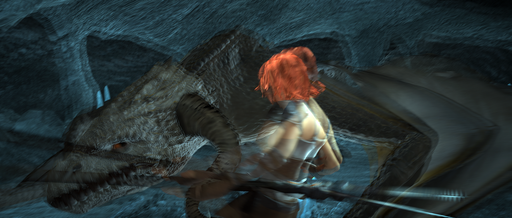}} &
  {\includegraphics[width=0.2\linewidth]{./resources/flow_examples/sintel.train.final_cave_4_49/gt.png}} &
  {\includegraphics[width=0.2\linewidth]{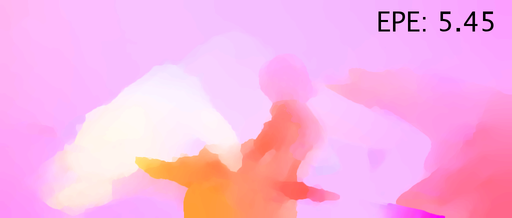}} &
  {\includegraphics[width=0.2\linewidth]{./resources/flow_examples/sintel.train.final_cave_4_49/dumbnet_43_chairs_moreds_540000_s=1.png}} &
  {\includegraphics[width=0.2\linewidth]{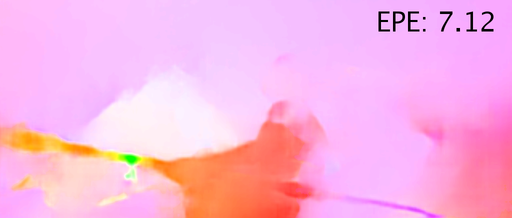}}
  \\
  {\includegraphics[width=0.2\linewidth]{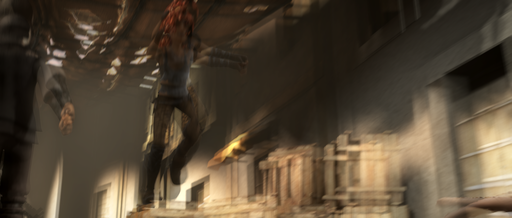}} &
  {\includegraphics[width=0.2\linewidth]{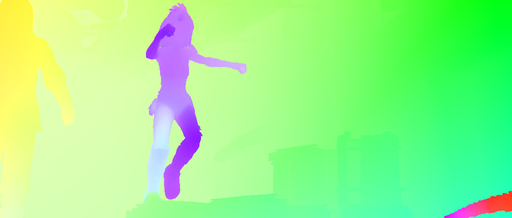}} &
  {\includegraphics[width=0.2\linewidth]{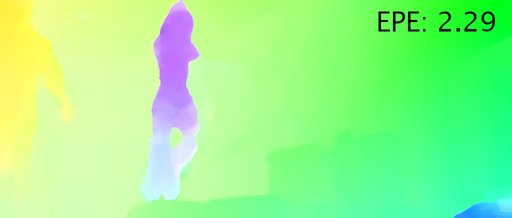}} &
  {\includegraphics[width=0.2\linewidth]{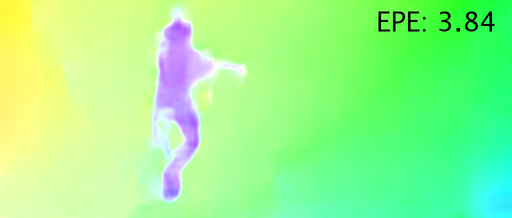}} &
  {\includegraphics[width=0.2\linewidth]{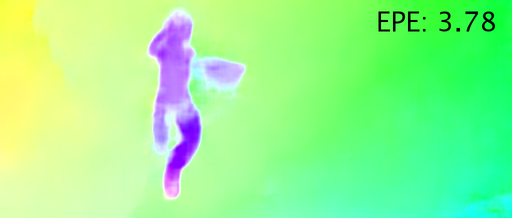}}
  \\
  {\includegraphics[width=0.2\linewidth]{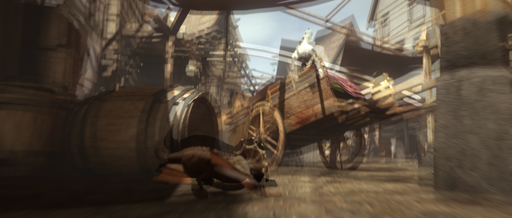}} &
  {\includegraphics[width=0.2\linewidth]{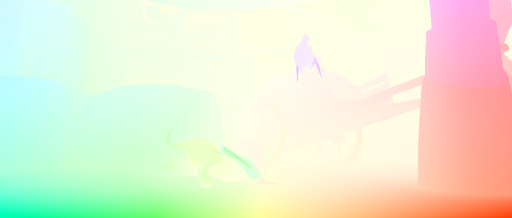}} &
  {\includegraphics[width=0.2\linewidth]{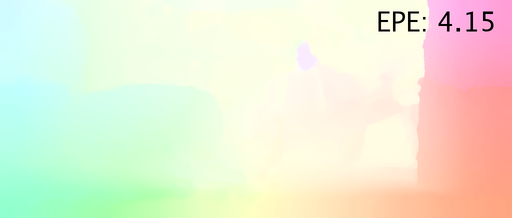}} &
  {\includegraphics[width=0.2\linewidth]{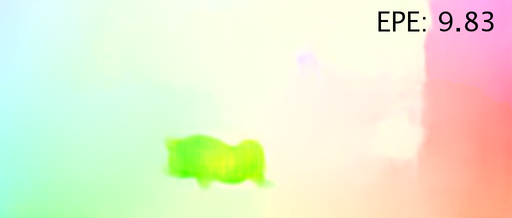}} &
  {\includegraphics[width=0.2\linewidth]{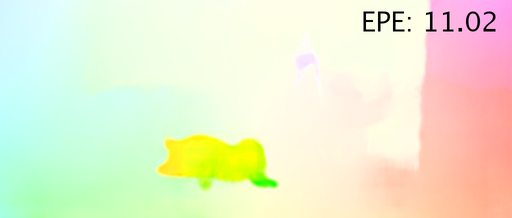}}
  \\
  {\includegraphics[width=0.2\linewidth]{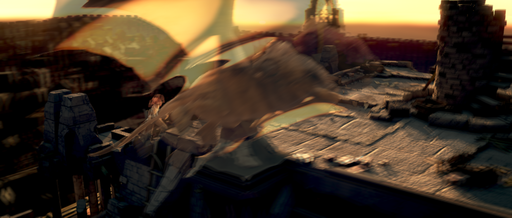}} &
  {\includegraphics[width=0.2\linewidth]{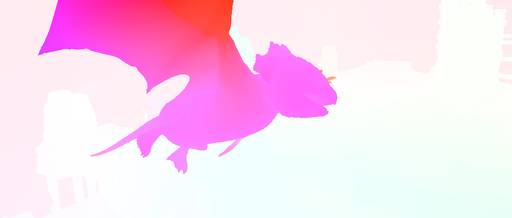}} &
  {\includegraphics[width=0.2\linewidth]{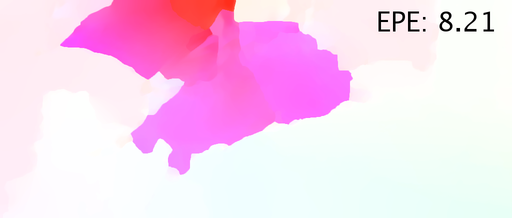}} &
  {\includegraphics[width=0.2\linewidth]{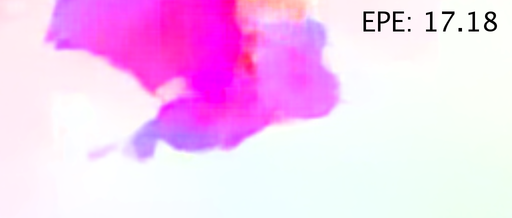}} &
  {\includegraphics[width=0.2\linewidth]{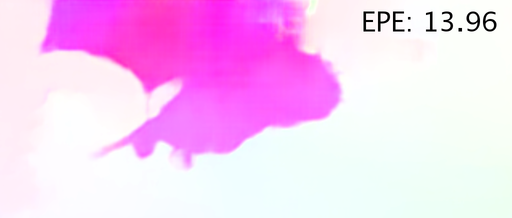}}
  \\
  {\includegraphics[width=0.2\linewidth]{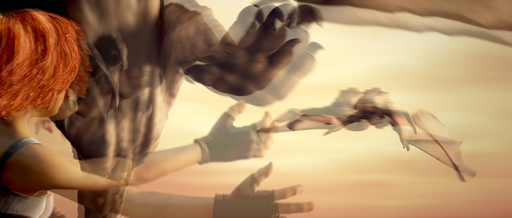}} &
  {\includegraphics[width=0.2\linewidth]{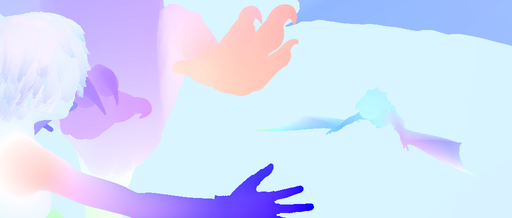}} &
  {\includegraphics[width=0.2\linewidth]{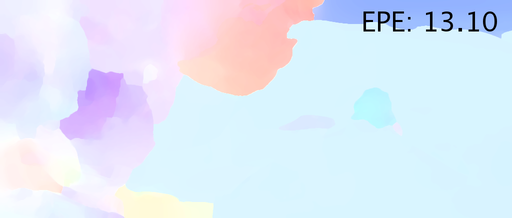}} &
  {\includegraphics[width=0.2\linewidth]{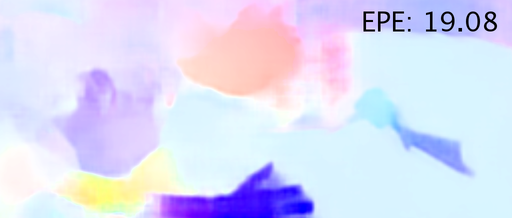}} &
  {\includegraphics[width=0.2\linewidth]{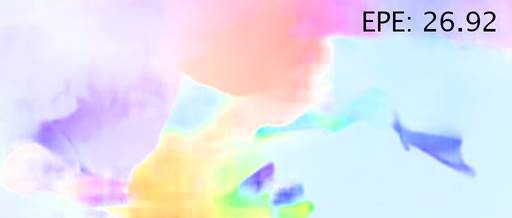}}

  \end{tabular}
\end{center}
   \caption{Examples of optical flow prediction on the Sintel dataset. In each row left to right: overlaid image pair, ground truth flow and 3 predictions:  EpicFlow, FlowNetS and FlowNetC. Endpoint error is shown for every frame. Note that even though the EPE of FlowNets is usually worse than that of EpicFlow, the networks often better preserve fine details.}
\label{fig:sintel_examples}
\end{figure*}

From Table~\ref{tbl:results_big} one can see that FlowNetC is better than FlowNetS on Sintel Clean, while on Sintel Final the situation changes. 
On this difficult dataset, FlowNetS+ft+v is even on par with DeepFlow.
Since the average endpoint error often favors over-smoothed solutions, it is interesting to see qualitative results of our method.
Figure~\ref{fig:sintel_examples} shows examples of the raw optical flow predicted by the two FlowNets (without fine-tuning), compared to ground truth and EpicFlow.
The figure shows how the nets often produce visually appealing results, but are still worse in terms of endpoint error.
Taking a closer look reveals that one reason for this may be the noisy non-smooth output of the nets especially in large smooth background regions.
This we can partially compensate with variational refinement.

\paragraph{KITTI.}
The KITTI dataset contains strong projective transformations which are very different from what the networks encountered during training on Flying Chairs.
Still, the raw network output is already fairly good, and additional fine-tuning and variational refinement give a further boost.
Interestingly, fine-tuning on Sintel improves the results on KITTI, probably because the images and motions in Sintel are more natural than in Flying Chairs.
The FlowNetS outperforms FlowNetC on this dataset.

\begin{figure}
\begin{center}
\setlength{\tabcolsep}{0.01cm}
\renewcommand{\arraystretch}{0.2}
  \begin{tabular}{ccccc}
  \small{Images} & \small{Ground truth} & \small{EpicFlow} & \small{FlowNetS} & \small{FlowNetC}
  \\
  {\includegraphics[width=0.2\linewidth]{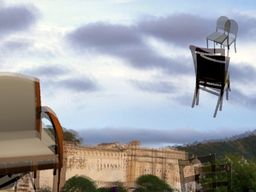}} &
  {\includegraphics[width=0.2\linewidth]{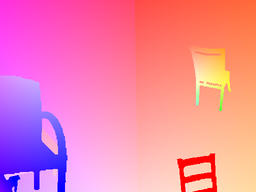}} &
  {\includegraphics[width=0.2\linewidth]{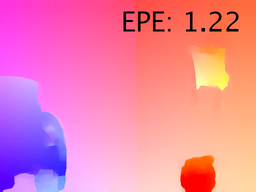}} &
  {\includegraphics[width=0.2\linewidth]{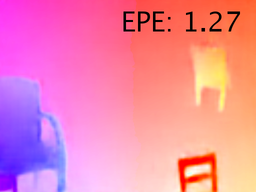}} &
  {\includegraphics[width=0.2\linewidth]{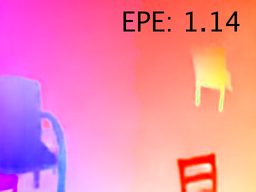}}
   \\
  {\includegraphics[width=0.2\linewidth]{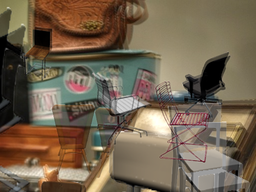}} &
  {\includegraphics[width=0.2\linewidth]{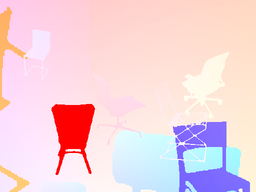}} &
  {\includegraphics[width=0.2\linewidth]{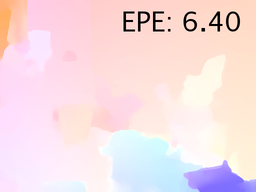}} &
  {\includegraphics[width=0.2\linewidth]{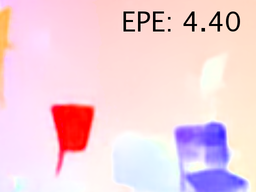}} &
  {\includegraphics[width=0.2\linewidth]{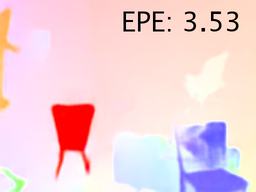}}
  \end{tabular}
\end{center}
   \caption{Examples of optical flow prediction on the Flying Chairs dataset. The images include fine details and small objects with large displacements which EpicFlow often fails to find. The networks are much more successful.}
\label{fig:chair_example_predictions}
\end{figure}

\paragraph{Flying Chairs.}
Our networks are trained on the Flying Chairs, and hence are expected to perform best on those.
When training, we leave aside a test set consisting of $640$ images.
Table~\ref{tbl:results_big} shows the results of various methods on this test set, some example predictions are shown in Fig.~\ref{fig:chair_example_predictions}.
One can see that FlowNetC outperforms FlowNetS and that the nets outperform all state-of-the-art methods.
Another interesting finding is that this is the only dataset where the variational refinement does not improve performance but makes things worse. Apparently the networks can do better than variational refinement already. This indicates that with a more realistic training set, the networks might also perform even better on other data.

\paragraph{Timings.}
In Table~\ref{tbl:results_big} we show the per-frame runtimes of different methods in seconds.
Unfortunately, many methods only provide the runtime on a single CPU, whereas our FlowNet uses layers only implemented on GPU.
While the error rates of the networks are below the state of the art, they are the best among real-time methods.
For both training and testing of the networks we use an \emph{NVIDIA GTX Titan} GPU. 
The CPU timings of DeepFlow and EpicFlow are taken from \cite{Revaud-et-al-15}, 
while the timing of LDOF was computed on a single 2.66GHz core.

\subsection{Analysis}\label{sec:analysis}
\paragraph{Training data.}
To check if we benefit from using the Flying Chairs dataset instead of Sintel, we trained a network just on Sintel, leaving aside a validation set to control the performance.
Thanks to aggressive data augmentation, even Sintel alone is enough to learn optical flow fairly well.
When testing on Sintel, the network trained exclusively on Sintel has EPE roughly $1$ pixel higher than the net trained on Flying Chairs and fine-tuned on Sintel.

The Flying Chairs dataset is fairly large, so is data augmentation still necessary? 
The answer is positive: training a network without data augmentation on the Flying Chairs results in an EPE increase of roughly $2$ pixels when testing on Sintel.

\paragraph{Comparing the architectures.}
The results in Table~\ref{tbl:results_big} allow to draw conclusions about strengths and weaknesses of the two architectures we tested.

First, FlowNetS generalizes to Sintel Final better than FlowNetC.
On the other hand, FlowNetC outperforms FlowNetS on Flying chairs and Sintel Clean.
Note that Flying Chairs do not include motion blur or fog, as in Sintel Final.
These results together suggest that even though the number of parameters of the two networks is virtually the same, the FlowNetC slightly more overfits to the training data.
This does not mean the network remembers the training samples by heart, but it adapts to the kind of data it is presented during training.
Though in our current setup this can be seen as a weakness, if better training data were available it could become an advantage.

Second, FlowNetC seems to have more problems with large displacements. This can be seen from the results on KITTI discussed above, and also from detailed performance analysis on Sintel Final (not shown in the tables). FlowNetS+ft achieves an s40+ error (EPE on pixels with displacements of at least 40 pixels) of $43.3$px, and for FlowNetC+ft this value is $48$px. One explanation is that the maximum displacement of the correlation does not allow to predict very large motions. This range can be increased at the cost of computational efficiency.

\section{Conclusion}\label{sec:conclusion}
Building on recent progress in design of convolutional network architectures, we have shown that it is possible to train a network to directly predict optical flow from two input images.
Intriguingly, the training data need not be realistic.
The artificial Flying Chairs dataset including just affine motions of synthetic rigid objects is sufficient to predict optical flow in natural scenes with competitive accuracy.
This proves the generalization capabilities of the presented networks.
On the test set of the Flying Chairs the CNNs even outperform state-of-the-art methods like DeepFlow and EpicFlow.
It will be interesting to see how future networks perform as more realistic training data becomes available.


\subsection*{Acknowledgments}
The work was partially funded by the ERC Starting Grants
VideoLearn and ConvexVision, by the DFG Grants BR-3815/7-1 and CR 250/13-1, and
by the EC FP7 project 610967 (TACMAN).

{\small
\bibliographystyle{ieee}
\bibliography{bibliography}

\begin{thebibliography}{10}\itemsep=-1pt

\bibitem{Aubry-et-al-14}
M.~Aubry, D.~Maturana, A.~Efros, B.~Russell, and J.~Sivic.
\newblock Seeing 3d chairs: exemplar part-based 2d-3d alignment using a large
  dataset of cad models.
\newblock In {\em CVPR}, 2014.

\bibitem{Baker-et-al-09}
S.~Baker, D.~Scharstein, J.~Lewis, S.~Roth, M.~J. Black, and R.~Szeliski.
\newblock A database and evaluation methodology for optical flow.
\newblock Technical Report MSR-TR-2009-179, December 2009.

\bibitem{Bao-et-al-14}
L.~Bao, Q.~Yang, and H.~Jin.
\newblock Fast edge-preserving patchmatch for large displacement optical flow.
\newblock In {\em CVPR}, 2014.

\bibitem{Black-97}
M.~J. Black, Y.~Yacoob, A.~D. Jepson, and D.~J. Fleet.
\newblock Learning parameterized models of image motion.
\newblock In {\em CVPR}, 1997.

\bibitem{BBPW04}
T.~Brox, A.~Bruhn, N.~Papenberg, and J.~Weickert.
\newblock High accuracy optical flow estimation based on a theory for warping.
\newblock In {\em ECCV}, volume 3024, pages 25--36. Springer, 2004.

\bibitem{Brox-Malik-11}
T.~Brox and J.~Malik.
\newblock Large displacement optical flow: descriptor matching in variational
  motion estimation.
\newblock {\em PAMI}, 33(3):500--513, 2011.

\bibitem{Butler-et-al-12}
D.~J. Butler, J.~Wulff, G.~B. Stanley, and M.~J. Black.
\newblock A naturalistic open source movie for optical flow evaluation.
\newblock In {A. Fitzgibbon et al. (Eds.)}, editor, {\em ECCV}, Part IV, LNCS
  7577, pages 611--625. Springer-Verlag, Oct. 2012.

\bibitem{Ciresan-et-al-12}
D.~C. Ciresan, L.~M. Gambardella, A.~Giusti, and J.~Schmidhuber.
\newblock Deep neural networks segment neuronal membranes in electron
  microscopy images.
\newblock In {\em NIPS}, pages 2852--2860, 2012.

\bibitem{Dosovitskiy-15}
A.~Dosovitskiy, J.~T. Springenberg, and T.~Brox.
\newblock Learning to generate chairs with convolutional neural networks.
\newblock In {\em CVPR}, 2015.

\bibitem{Eigen-et-al-14}
D.~Eigen, C.~Puhrsch, and R.~Fergus.
\newblock Depth map prediction from a single image using a multi-scale deep
  network.
\newblock {\em NIPS}, 2014.

\bibitem{Farabet-13}
C.~Farabet, C.~Couprie, L.~Najman, and Y.~LeCun.
\newblock Learning hierarchical features for scene labeling.
\newblock {\em PAMI}, 35(8):1915--1929, 2013.

\bibitem{Fischer-14}
P.~Fischer, A.~Dosovitskiy, and T.~Brox.
\newblock Descriptor matching with convolutional neural networks: a comparison
  to {SIFT}.
\newblock 2014.
\newblock pre-print, arXiv:1405.5769v1 [cs.CV].

\bibitem{Ganin-et-al-14}
Y.~Ganin and V.~S. Lempitsky.
\newblock N{\^{}}4 -fields: Neural network nearest neighbor fields for image
  transforms.
\newblock In {\em ACCV}, pages 536--551, 2014.

\bibitem{Geiger-et-al-13}
A.~Geiger, P.~Lenz, C.~Stiller, and R.~Urtasun.
\newblock Vision meets robotics: The kitti dataset.
\newblock {\em International Journal of Robotics Research (IJRR)}, 2013.

\bibitem{Girshick-et-al-14}
R.~Girshick, J.~Donahue, T.~Darrell, and J.~Malik.
\newblock Rich feature hierarchies for accurate object detection and semantic
  segmentation.
\newblock In {\em CVPR}, 2014.

\bibitem{Goodfellow-14}
I.~Goodfellow, J.~Pouget-Abadie, M.~Mirza, B.~Xu, D.~Warde-Farley, S.~Ozair,
  A.~Courville, and Y.~Bengio.
\newblock Generative adversarial nets.
\newblock In {\em NIPS}, 2014.

\bibitem{Hariharan-et-al-15}
B.~Hariharan, P.~Arbeláez, R.~Girshick, and J.~Malik.
\newblock Hypercolumns for object segmentation and fine-grained localization.
\newblock {\em CVPR}, 2015.

\bibitem{Hays-Efros-08}
J.~Hays and A.~A. Efros.
\newblock im2gps: estimating geographic information from a single image.
\newblock In {\em CVPR}, 2008.

\bibitem{Horn-Schunck-81}
B.~K.~P. Horn and B.~G. Schunck.
\newblock Determining optical flow.
\newblock {\em Artificial Intelligence}, 17:185--203, 1981.

\bibitem{caffe}
Y.~Jia, E.~Shelhamer, J.~Donahue, S.~Karayev, J.~Long, R.~Girshick,
  S.~Guadarrama, and T.~Darrell.
\newblock Caffe: Convolutional architecture for fast feature embedding.
\newblock {\em arXiv preprint arXiv:1408.5093}, 2014.

\bibitem{Kennedy-et-al-15}
R.~Kennedy and C.~Taylor.
\newblock Optical flow with geometric occlusion estimation and fusion of
  multiple frames.
\newblock In {\em EMMCVPR}. 2015.

\bibitem{Kingma-14}
D.~P. Kingma and J.~Ba.
\newblock Adam: {A} method for stochastic optimization.
\newblock In {\em ICLR}, 2015.

\bibitem{Konda-Memisevic-13}
K.~R. Konda and R.~Memisevic.
\newblock Unsupervised learning of depth and motion.
\newblock {\em CoRR}, abs/1312.3429, 2013.

\bibitem{Krizhevsky-et-al-12}
A.~Krizhevsky, I.~Sutskever, and G.~E. Hinton.
\newblock Imagenet classification with deep convolutional neural networks.
\newblock In {\em NIPS}, pages 1106--1114, 2012.

\bibitem{lecun1989backpropagation}
Y.~LeCun, B.~Boser, J.~S. Denker, D.~Henderson, R.~E. Howard, W.~Hubbard, and
  L.~D. Jackel.
\newblock Backpropagation applied to handwritten zip code recognition.
\newblock {\em Neural computation}, 1(4):541--551, 1989.

\bibitem{Leordeanu-et-al-12}
M.~Leordeanu, R.~Sukthankar, and C.~Sminchisescu.
\newblock Efficient closed-form solution to generalized boundary detection.
\newblock In {\em Proceedings of the 12th European Conference on Computer
  Vision - Volume Part IV}, ECCV'12, pages 516--529, Berlin, Heidelberg, 2012.
  Springer-Verlag.

\bibitem{Leordeanu-et-al-13}
M.~Leordeanu, A.~Zanfir, and C.~Sminchisescu.
\newblock Locally affine sparse-to-dense matching for motion and occlusion
  estimation.
\newblock {\em ICCV}, 0:1721--1728, 2013.

\bibitem{Long-et-al-15}
J.~Long, E.~Shelhamer, and T.~Darrell.
\newblock Fully convolutional networks for semantic segmentation.
\newblock In {\em CVPR}, 2015.

\bibitem{MP98}
E.~M\'emin and P.~P\'erez.
\newblock Dense estimation and object-based segmentation of the optical flow
  with robust techniques.
\newblock 7(5):703--719, May 1998.

\bibitem{Revaud-et-al-15}
J.~Revaud, P.~Weinzaepfel, Z.~Harchaoui, and C.~Schmid.
\newblock {EpicFlow: Edge-Preserving Interpolation of Correspondences for
  Optical Flow}.
\newblock In {\em CVPR}, Boston, United States, June 2015.

\bibitem{Rosenbaum-13}
D.~Rosenbaum, D.~Zoran, and Y.~Weiss.
\newblock Learning the local statistics of optical flow.
\newblock In {\em NIPS}, 2013.

\bibitem{Sun-08}
D.~Sun, S.~Roth, J.~Lewis, and M.~J. Black.
\newblock Learning optical flow.
\newblock In {\em ECCV}, 2008.

\bibitem{Taylor-10}
G.~W. Taylor, R.~Fergus, Y.~LeCun, and C.~Bregler.
\newblock Convolutional learning of spatio-temporal features.
\newblock In {\em ECCV}, pages 140--153, 2010.

\bibitem{Wedel-et-al-iccv09}
A.~Wedel, D.~Cremers, T.~Pock, and H.~Bischof.
\newblock Structure- and motion-adaptive regularization for high accuracy optic
  flow.
\newblock In {\em ICCV}, Kyoto, Japan, 2009.

\bibitem{Weinzaepfelet-al-13}
P.~Weinzaepfel, J.~Revaud, Z.~Harchaoui, and C.~Schmid.
\newblock {DeepFlow: Large displacement optical flow with deep matching}.
\newblock In {\em ICCV}, Sydney, Australia, Dec. 2013.

\bibitem{Zbontar-14}
J.~Zbontar and Y.~LeCun.
\newblock Computing the stereo matching cost with a convolutional neural
  network.
\newblock {\em CoRR}, abs/1409.4326, 2014.

\bibitem{Zeiler-14}
M.~D. Zeiler and R.~Fergus.
\newblock Visualizing and understanding convolutional networks.
\newblock In {\em ECCV}, 2014.

\bibitem{Zeiler-11}
M.~D. Zeiler, G.~W. Taylor, and R.~Fergus.
\newblock Adaptive deconvolutional networks for mid and high level feature
  learning.
\newblock In {\em ICCV}, pages 2018--2025, 2011.

\end{thebibliography}
}

\clearpage
\twocolumn[
  \begin{@twocolumnfalse}
{
   \newpage
   \null
   \vskip .375in
   \begin{center}
      {\Large \bf Supplementary Material for `FlowNet: Learning Optical Flow with Convolutional Networks' \par}
      \vspace*{24pt}
      {
      \large
      \lineskip .5em
      \begin{tabular}[t]{c}
          
      \end{tabular}
      \par
      }
      \vskip .5em
      \vspace*{12pt}
   \end{center}
}
  \end{@twocolumnfalse}
]
\setcounter{section}{0}
\setcounter{figure}{0}
\setcounter{table}{0}
\setcounter{footnote}{0}

\renewcommand*{\theHsection}{A\thesection}
\renewcommand*{\theHfigure}{A\thefigure}
\renewcommand*{\theHtable}{A\thetable}

\section{Flow field color coding}
\begin{figure}
\begin{center}
\includegraphics[width=0.5\linewidth]{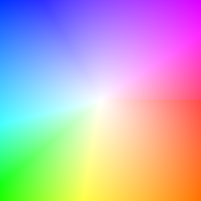}
\end{center}
\caption{Flow field color coding. The central pixel does not move, and the displacement of every other pixel is the vector from the center to this pixel.}
\label{fig:flow_key}
\end{figure}

To visualize the flow fields, we use the tool provided with Sintel~\cite{Butler-et-al-12}. 
Flow direction is encoded with color and magnitude with color intensity. 
White corresponds to no motion. 
Figure~\ref{fig:flow_key} illustrates flow color coding: the flow vector at each pixel is a vector from the center of the square to this pixel.
Since the magnitudes of flows in different image pairs shown in the main paper are very different, we independently normalize the maximum color intensity for each image pair, but in the same way for different methods applied to one image pair.

\section{Details of generating Flying Chairs}\label{sec:detchairs}
We explain in detail the process of generating the Flying Chairs dataset.
As background we use $964$ images of resolution $1024 \times 768$ pixels, downloaded from Flickr. 
As foreground objects we use $809$ chair models from the dataset of Aubry \etal~\cite{Aubry-et-al-14}, each rendered from $62$ views: $31$ azimuth angles and $2$ elevation angles.
To generate the first image in an image pair, we take a background image and randomly position a random set of chairs ontop.
The number of the chairs is sampled uniformly from $[16;\, 24]$,
the types and viewpoints of the chairs are sampled uniformly
and the locations of the chairs are sampled uniformly from the whole image.
The sizes of the chairs (in pixels) are sampled from a Gaussian with mean $200$ and standard deviation $200$, and then clamped between $50$ and $640$.

\begin{figure}
\begin{center}
\begin{tabular}{cc}
\includegraphics[width=0.48\linewidth]{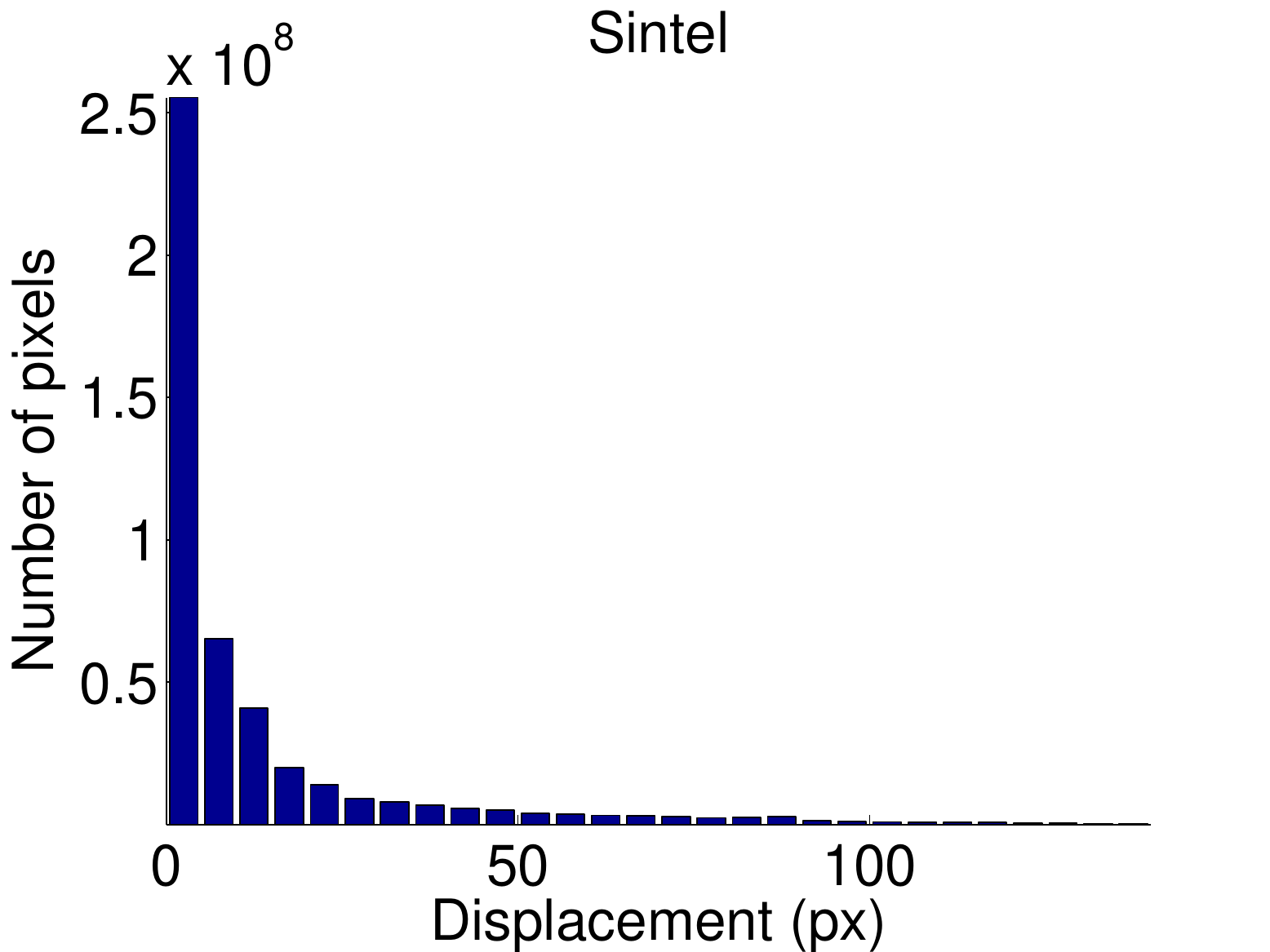}&
\includegraphics[width=0.48\linewidth]{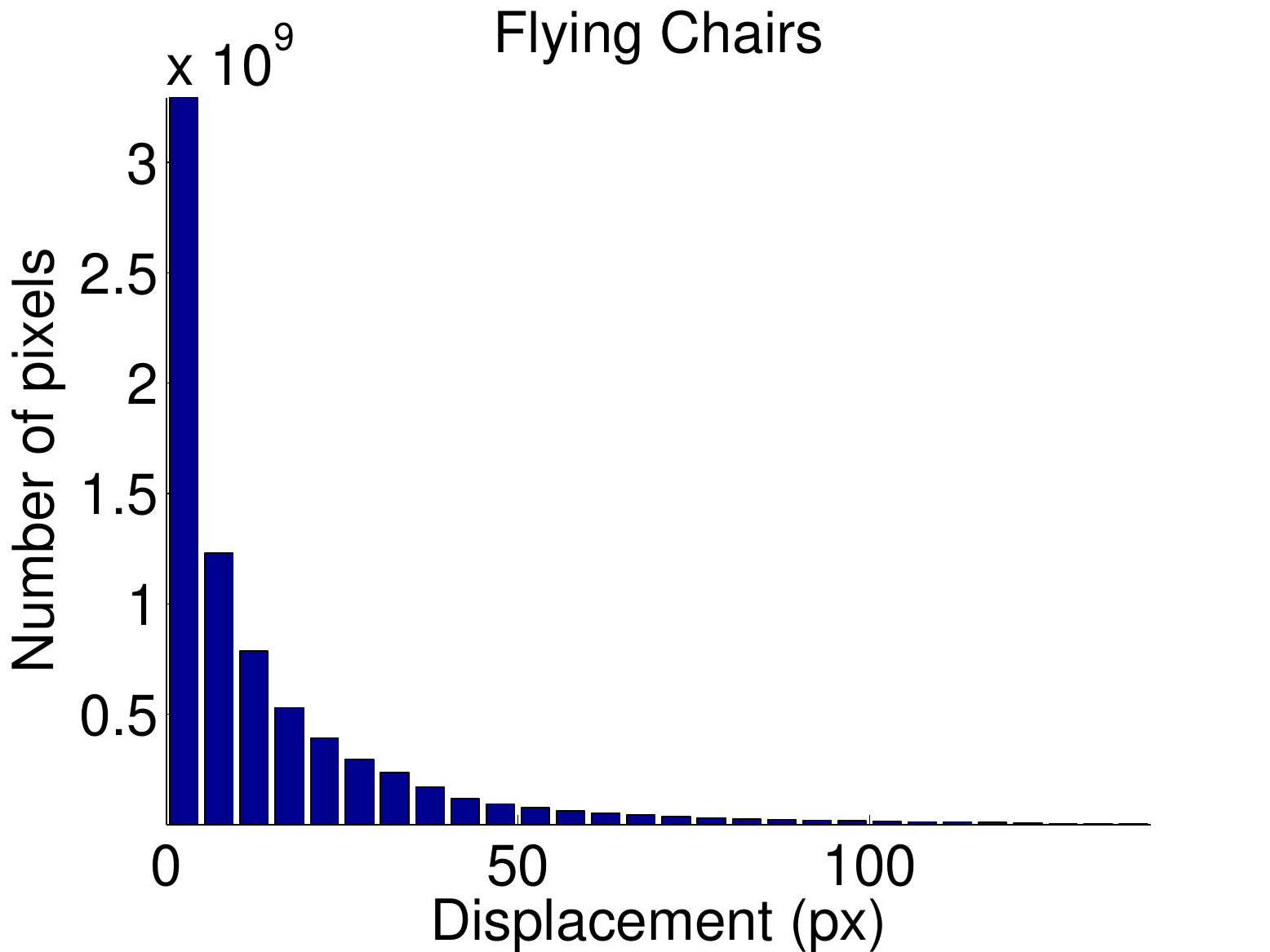}\\
\includegraphics[width=0.48\linewidth]{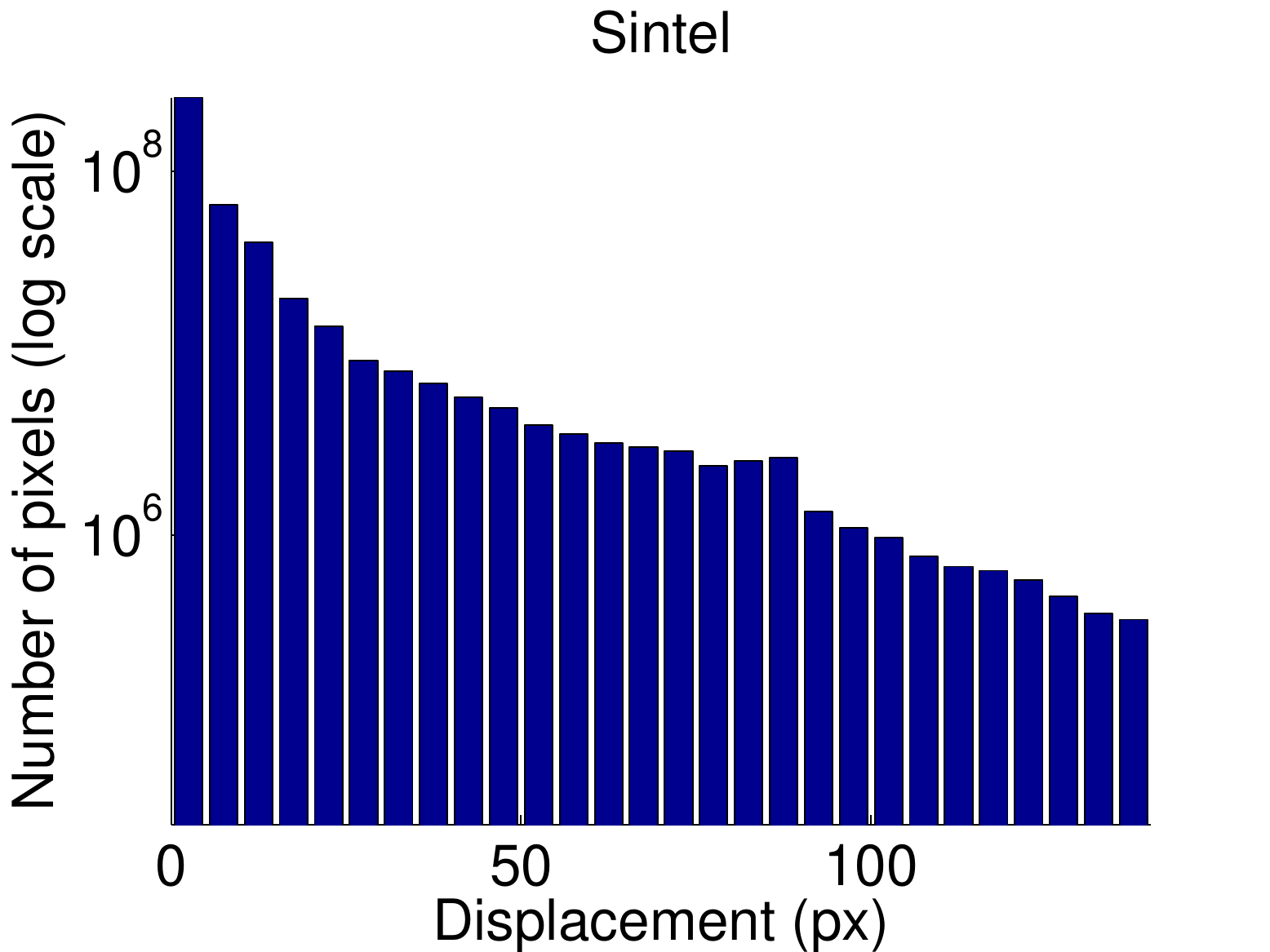} &
\includegraphics[width=0.48\linewidth]{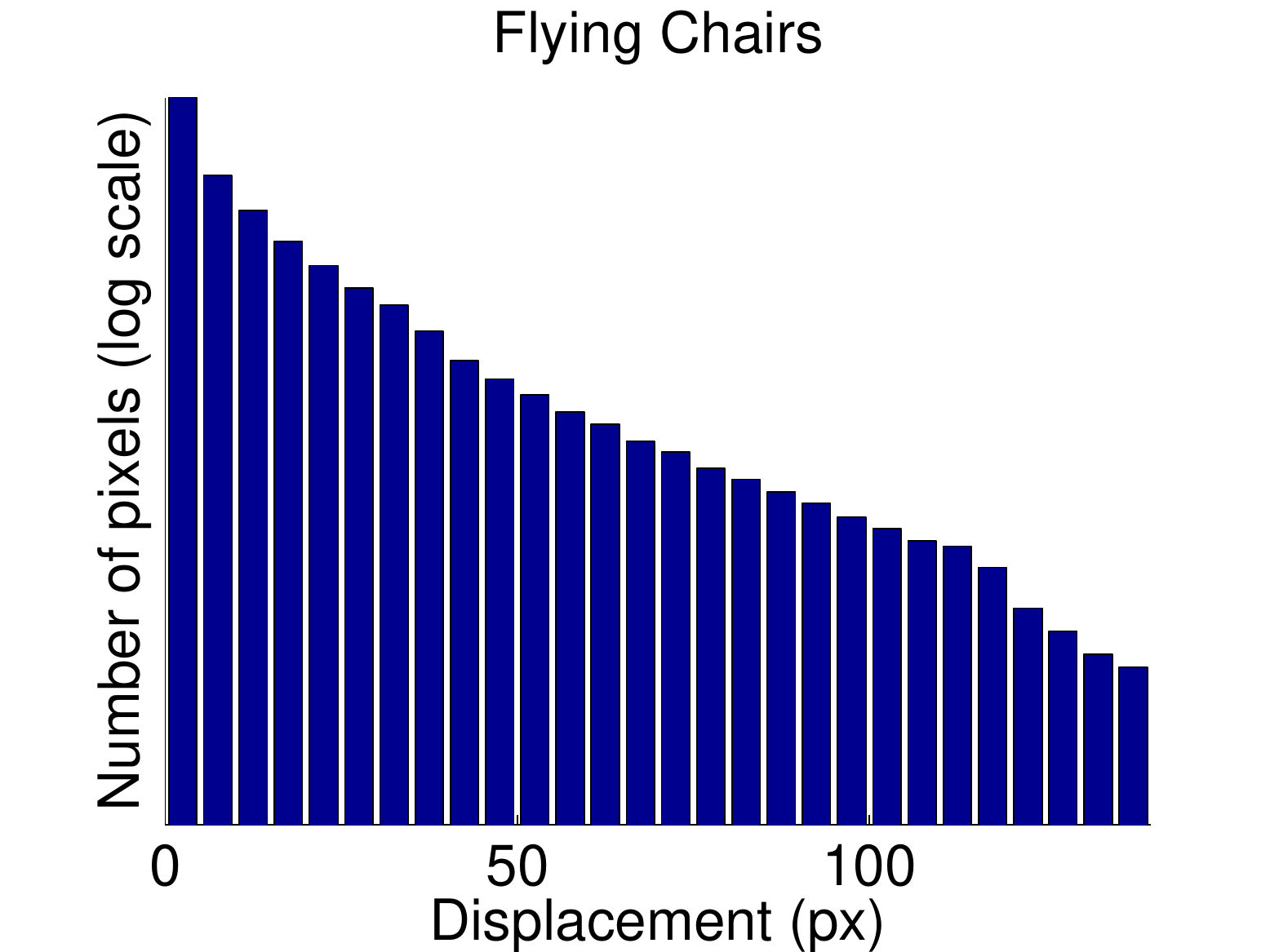}
\end{tabular}
\end{center}
\caption{Histogram of displacement distribution in Sintel (left) and Flying Chairs (right) with linear (top) and logarithmic (bottom) y-axis. The distribution was cut off at the displacement of 150 pixels, the maximum flow in Sintel is actually around 450 pixels.}
\label{fig:displacement-hist}
\end{figure}

To generate the second image in a pair and the flow field, we apply random transformations to the chairs and the background.
Each of these transformations is a composition of zooming, rotation and translation.
The parameters to sample are the zoom coefficient, the rotation angle and the translation vector.
We aim to roughly match the displacement distribution of Sintel, shown in Fig.~\ref{fig:displacement-hist} (left).
Simply sampling the transformation parameters from Gaussians results in too few small displacements,
we hence make the distributions of the transformation parameters to be more peaked around zero than Gaussians.

The family of distributions from which we sample the parameters 
contains mixtures of two distributions: a constant $\mu$ with probability $1-p$ and a power of a Gaussian with probability $p$.
More precisely, let $\gamma \sim \mathcal{N} (\mu,\, \sigma)$ be a univariate Gaussian. 
We raise its absolute value to a power $k$ (keeping the sign) and clamp to the interval $[a,\,b]$. 
We then set the value to $\mu$ with probability $1-p$.
Overall, the result is given by:
$$\xi = \beta \cdot \max(\min(\sign(\gamma)\cdot |\gamma|^k ,b),a) + (1-\beta) \cdot \mu,$$
where $\beta$ is a Bernoulli random variable equaling $1$ with probability $p$ and $0$ with probability $1-p$.
We denote the distribution of $\xi$ by $G(k, \mu,\, \sigma,\, a,\, b,\, p)$.
All transformation parameters are sampled from distributions from this family, with parameters shown in Table~\ref{tbl:transf_param_distrib}.

Given the transformation parameters, it is straightforward to generate the second image in the pair, as well as the flow field and the occlusion map. 
We then cut each image into $4$ quarters, resulting in $4$ image pairs of size $512 \times 384$ pixels each. 
The displacement histogram of the Flying Chairs dataset is shown in Fig.~\ref{fig:displacement-hist} (right).

\begin{table}
\begin{tabular}{c|rrrrrr}
Parameter       & $k$ & $\mu$ & $\sigma$ &  $a$   &  $b$   &  $p$   \\ \hline
Translation BG  & $4$ &  $0$  &  $1.3$   & $-40$  & $40$   &  $1$   \\
Rotation BG     & $2$ &  $0$  &  $1.3$   & $-10$  &  $10$  & $0.3$  \\
Zoom BG         & $2$ &  $1$  &  $0.1$   & $0.93$ & $1.07$ & $0.6$  \\
Translation CH  & $3$ &  $0$  &  $2.3$   & $-120$ & $120$  &  $1$   \\
Rotation CH     & $2$ &  $0$  &  $2.3$   & $-30$  & $30$   & $0.7$  \\
Zoom CH         & $2$ &  $1$  &  $0.18$  & $0.8$  & $1.2$  & $0.7$  \\
\end{tabular}
\vspace*{0.2cm}
\caption{Parameters of the distribution of the transformation parameters.}
\label{tbl:transf_param_distrib}
\end{table}

We did not study in detail the effect of the dataset parameters on the FlowNet results. However, we observed, that with much simpler strategy of sampling all transformation parameters from Gaussians the networks still work, but are less accurate than networks trained on the data described above.

\begin{figure}[b]
\begin{center}
\includegraphics[width=0.85\linewidth]{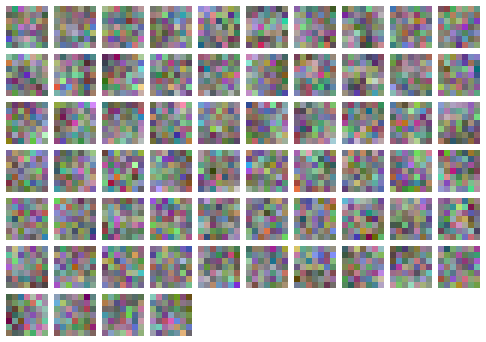}
\end{center}
\caption{First layer filters of FlowNetCorr. The filters are noisy, but some structure is still visible.}
\label{fig:corr_filters_conv1}
\end{figure}

\section{Convolutional Filters}\label{sec:filters}
\begin{figure*}
\begin{center}
\includegraphics[width=0.75\linewidth]{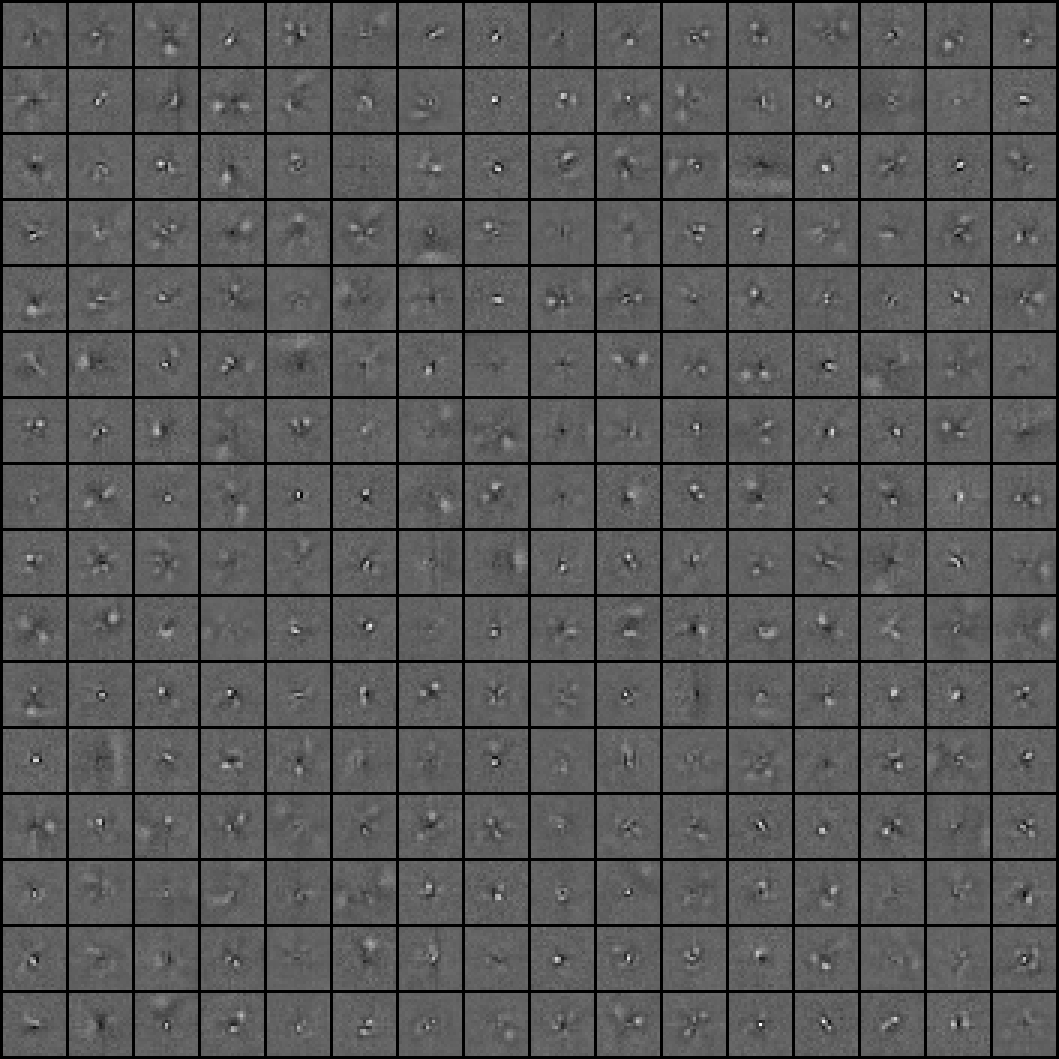}
\end{center}
\caption{Visualization of filters applied on top of the correlation layer in FlowNetCorr. There are $256$ filters and for each of them we show the weights shaped as a $21 \times 21$ pixels patch, where each pixel corresponds to a displacement vector. The center pixel of each patch corresponds to zero displacement. The filters favor interesting unique displacement patterns.}
\label{fig:corr_filters}
\end{figure*}

When taking a closer look at the filters of the FlowNets, one can see that lower layer filters have few structure and higher layer filters are more structured. Fig.~\ref{fig:corr_filters_conv1} shows how the first layer filters have not completely converged, however coarse gradients are visible. In contrast, the filters that are applied to the output of the correlation layer have very visible structure, as shown in Fig.~\ref{fig:corr_filters}. Different filters are selective for different flow directions and magnitudes.

\begin{figure}
\begin{center}
\includegraphics[width=0.95\linewidth]{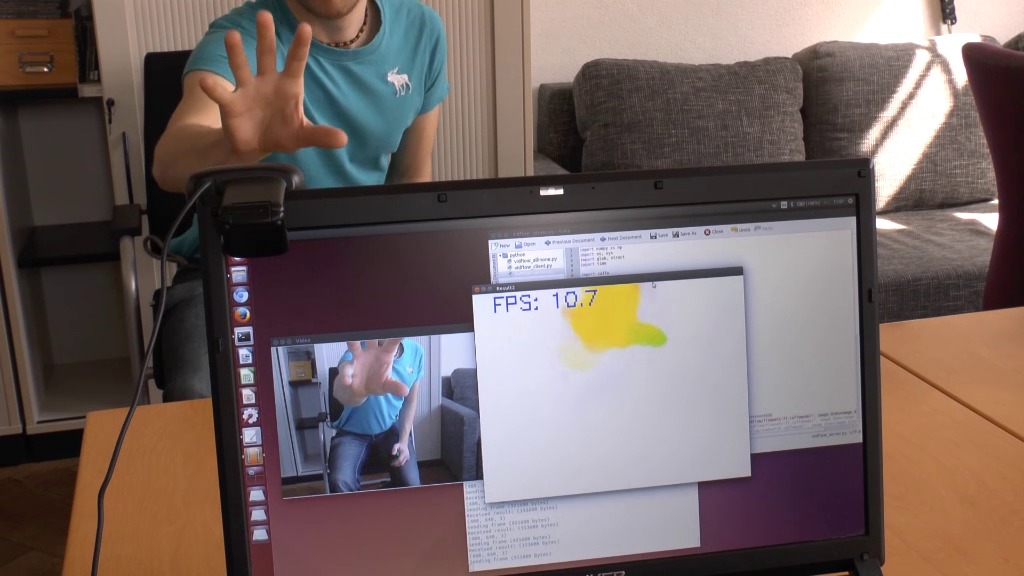}
\end{center}
\caption{Demo Video: Application of FlowNets to a live video stream. Watch on YouTube: \href{http://goo.gl/YmMOkR}{http://goo.gl/YmMOkR}}
\label{fig:demovid}
\end{figure}

\section{Video}
In the supplementary video we demonstrate the real-time operation of the FlowNets using a notebook with a GeForce GTX 980M GPU. 
Resolution of images captured with a \mbox{webcam} is $640 \times 480$ pixels.
We show example flow fields produced from real-life videos by both FlowNetSimple and FlowNetCorr for indoor and outdoor scenes.
The video can be found on \href{http://goo.gl/YmMOkR}{http://goo.gl/YmMOkR}.
A sample frame from the video can be seen in Fig.~\ref{fig:demovid}.

\newpage

\end{document}